\documentclass{article}

\usepackage[preprint]{neurips_2026}

\usepackage[utf8]{inputenc} % allow utf-8 input
\usepackage[T1]{fontenc}    % use 8-bit T1 fonts
\usepackage{hyperref}       % hyperlinks
\usepackage{url}            % simple URL typesetting
\usepackage{booktabs}       % professional-quality tables
\usepackage{amsfonts}       % blackboard math symbols
\usepackage{nicefrac}       % compact symbols for 1/2, etc.
\usepackage{microtype}      % microtypography
\usepackage{xcolor}         % colors

\usepackage{amsmath,amssymb,amsthm}
\usepackage{booktabs}
\usepackage{hyperref}
\usepackage{cleveref}
\usepackage{algorithm}
\usepackage{algorithmic}
\usepackage{graphicx}
\usepackage{xcolor}
\usepackage{enumitem}
\usepackage{multirow}
\usepackage{microtype}
\usepackage{natbib}

\theoremstyle{remark}
\newtheorem{remark}{Remark}

\usepackage{subcaption}
\usepackage{wrapfig}

\title{Target-Aware Data Augmentation for SAT Prediction}

\author{%
  Eshed Gal \\
  Department of Computer Science\\
  The University of British Columbia\\
  Vancouver, Canada \\
  \texttt{eshedg@cs.ubc.ca} \\
  \And
  Uri Ascher \\
  Department of Computer Science\\
  The University of British Columbia\\
  Vancouver, Canada \\
  \texttt{ascher@cs.ubc.ca} \\
  \And
  Eldad Haber \\
  Department of Earth, Ocean and Atmospheric Sciences\\
  The University of British Columbia\\
  Vancouver, Canada \\
  \texttt{ehaber@eoas.ubc.ca} \\
}

\begin{document}

\maketitle

\begin{abstract}
Learning-based approaches to NP-hard problems have shown increasing promise, but their progress is fundamentally constrained by the high cost of generating labeled training data. In domains such as Boolean satisfiability (SAT), standard pipelines rely on solver-in-the-loop labeling, which scales poorly with problem size and limits the amount of usable supervision. This bottleneck hinders the broader goal of leveraging machine learning to capture structure in hard combinatorial problems.
In this work, we propose a target-aware, solver-free data generation framework for SAT that produces correctly labeled SAT and UNSAT instances by construction, eliminating the need for expensive solver calls. Our method aligns generated instances with the structural properties of a target benchmark, making synthetic data effective for downstream learning.
We further develop a linear-programming-aware graph neural network (LPGNN) architecture that incorporates constraint-violation residuals into message passing, enabling the model to exploit underlying optimization structure. Together, these contributions support a data-centric paradigm for learning on NP-hard problems, where scalable, task-aligned data generation is as critical as model design. Our approach yields orders-of-magnitude speedups in data generation, demonstrating that benchmark-aligned synthetic data can effectively augment solver-labeled datasets for GNN-based SAT prediction.
\end{abstract}

\section{Introduction}

Boolean satisfiability (SAT) is a canonical NP-complete problem \citep{cook1971}. This is a core primitive in formal verification \citep{Biereetal1999}, combinatorial optimization \citep{Moskewiczetal2001}, and cryptanalysis \citep{Soosetal2009}. Despite the worst-case NP-hardness of SAT, modern conflict-driven clause-learning (CDCL) solvers such as CaDiCaL \citep{biere2020cadical} can approximately solve many large practical instances efficiently. At the same time, solver performance remains highly sensitive to instance structure, and the design of effective heuristics remains difficult and domain-specific. This has motivated increasing interest in machine learning methods for SAT, with the broader goal of using learned structure to improve performance on hard combinatorial problems.

% Recent work has shown that Graph Neural Networks (GNNs) can learn useful representations of SAT formulas from graph-based encodings. NeuroSAT \citep{selsam2019} demonstrated that satisfiability can be predicted from single-bit supervision, and subsequent works studied alternative graph constructions, learned search guidance, and optimization-aware architectures \citep{li2024g4satbench,cheng2024satgl,ansotegui2012,neuroback2024,yolcu2019,peltonen2026,eliasof2024,khalil2022mipgnn,cardillo2025}. Together with benchmark suites such as G4SATBench \citep{li2024g4satbench}, these advances have established SAT as a useful testbed for learning on NP-hard problems.

Recent works have shown that Graph Neural Networks (GNNs) can learn useful representations of SAT formulas from graph-based encodings. 
% NeuroSAT \citep{selsam2019} demonstrated that satisfiability can be predicted from single-bit supervision, and subsequent research has diversified into several specialized directions. 
One major thread explores alternative graph constructions and literal-incidence encodings to better capture the underlying community and algebraic structure of formulas \citep{ansotegui2012, cheng2024satgl, peltonen2026}. Another focuses on providing learned search guidance, ranging from local search heuristics to improving CDCL solver decisions \citep{yolcu2019, neuroback2024}. Some works focus on optimization-aware architectures that incorporate algebraic feasibility signals, such as linear programming residuals or energy-based functions, directly into message passing \citep{khalil2022mipgnn, eliasof2024, cardillo2025}. Together with comprehensive benchmark suites such as G4SATBench \citep{li2024g4satbench}, these advances have established SAT as a primary testbed for scaling machine learning methods on NP-hard problems.

A basic premise of machine learning is that performance depends jointly on the architecture, the size of the model, and the quality and scale of the training data. The recent history of language modeling illustrates this point: progress accelerated once next-token prediction provided a scalable self-supervised objective, allowing increasingly large architectures to be trained on massive corpora \citep{kaplan2020scaling,hoffmann2022training,radford2019language,brown2020language}. Without a scalable source of supervision, however, architectural improvements alone are difficult to exploit.

For SAT learning, recent work has focused on graph representations and GNN architectures, and we also introduce architectural improvements below. However, SAT does not naturally come with a next-token-prediction analogue: supervision is typically obtained by generating formulas and then calling a complete SAT solver to determine whether each instance is SAT (satisfiable formula) or UNSAT (unsatisfiable formula). As formulas become larger and harder, this solver-in-the-loop labeling process becomes increasingly costly, limiting the amount of usable supervision and tying learning progress to exact solver runtime. Recent works have begun to highlight this issue explicitly \citep{selfsatisfied2024}, and we discuss additional related literature in \Cref{app:related_work}.

In this work, we address this data bottleneck directly. We propose a target-aware, solver-free data-generation framework that constructs SAT and UNSAT formulas with certified labels by construction while matching structural properties of a target benchmark. The goal is not merely to generate more data, but to generate data that is useful for downstream learning. In particular, we show that synthetic SAT data can be generated with meaningful benchmark-aligned structure. Such data, when combined with GNN architectures, improves predictive performance. This supports a data-centric view: progress depends not only on better architectures, but also on scalable mechanisms for producing informative training data.
Although synthetic data augmentation carries an inherent risk of introducing distributional bias, prior work suggests it can act as an effective regularizer, often improving model robustness and downstream generalization \citep{dao2019kernel, shorten2019survey}.

The starting point of our approach is a simple asymmetry. While solving SAT is NP-hard, that is, finding a satisfying assignment for a given formula is difficult in the worst case; constructing a formula around a known assignment is easier. This is analogous in spirit to the method of manufactured solutions in numerical partial differential equations, where one first chooses an exact solution and then constructs a problem for which that solution is known \citep{salari2000code}. We exploit the same inverse viewpoint for SAT data generation: rather than sampling a formula and then invoking a solver to determine its label, we first construct a certificate, a satisfying assignment for SAT instances or an explicit contradiction for UNSAT instances, and then generate a CNF formula consistent with that certificate.

Our generation method is therefore CNF-native and solver-free. For SAT instances, we construct clauses around a planted assignment. For UNSAT instances, we embed a certified contradiction and add benchmark-aligned filler clauses. In both cases, the label is known by construction, avoiding repeated calls to a complete solver during data generation. At the same time, target-aware matching reduces the distributional mismatch that would arise from generic planted constructions.

In addition, we connect this data pipeline to an optimization-aware model design. Using the standard binary linear encoding of CNF formulas, we represent SAT as a feasibility problem and derive a Linear Programming (LP)-aware residual mechanism that injects constraint-violation information into message passing. This exposes algebraic feasibility structure to the GNN and complements the proposed generation framework, following the general direction of optimization-aware graph learning \citep{eliasof2024,khalil2022mipgnn}. Overall, this work advances a simple claim: for machine learning on NP-hard problems, scalable and task-aligned data generation can be as important as model architecture.

\paragraph{Contributions.}
\begin{itemize}
    \item We propose a \textbf{target-aware solver-free data-generation framework} for SAT learning, motivated by the broader challenge of scaling machine learning methods for NP-hard problems.

    \item We introduce a \textbf{CNF-native formula-generation strategy} that produces correctly labeled SAT and UNSAT formulas by construction, avoiding repetitive solver calls during data generation.

    \item We develop a \textbf{benchmark-aligned augmentation approach} that matches generated formulas to structural and algebraic properties of a target distribution, making synthetic data more useful for downstream learning than generic generated instances.

    \item We formulate the learning pipeline through a \textbf{binary linear feasibility view} that connects CNF formulas, generated instances, and graph-based learning in a common algebraic framework.

    \item We introduce an \textbf{LP-aware residual GNN architecture} that incorporates feasibility-related signals into message passing, complementing the proposed data-generation framework with an optimization-aware model component.

    % \item We provide empirical evidence that \textbf{data scale and model scale interact}: target-aware synthetic data enables larger GNNs to achieve improved predictive performance, with trends consistent with empirical scaling behavior.
\end{itemize}

\section{Related Work}

\paragraph{Graph representations for SAT.}
A central modeling decision in SAT learning is how a CNF formula is represented as a graph. Bipartite constructions dominate the recent literature. NeuroSAT \citep{selsam2019} and G4SATBench represent formulas using literal--clause graphs, whereas MILP-SAT-GNN \citep{cardillo2025} builds a variable-constraint graph from a mixed-integer formulation. These encodings preserve clause structure and expose the incidence pattern between variables and constraints, but they also imply that variables co-occurring in a clause remain separated by at least two hops in the message-passing graph. Alternative non-bipartite constructions include the variable interaction graph (VIG) and related variants \citep{ansotegui2012}. Empirical comparisons, however, generally favor bipartite encodings for satisfiability prediction: SATGL \citep{cheng2024satgl} reports a clear advantage for literal--clause graphs, and \citet{fu2025structuresat} similarly find non-bipartite structures to be substantially weaker for this task.

\textbf{Data augmentation in machine learning.} Beyond SAT-specific formula generation, our work connects to the broader paradigm of data augmentation, which is a cornerstone of robust representation learning \citep{shorten2019survey, cubuk2020randaugment}. While continuous domains such as computer vision readily admit semantics-preserving perturbations (e.g., cropping, rotation), augmenting discrete, graph-structured data is inherently more challenging \citep{ding2022data, zhao2022graph}. Graph data augmentation techniques often rely on random node dropping, feature masking, or edge perturbation \citep{rong2020dropedge, you2020graph}. However, applying these generic perturbations to combinatorial problems like SAT is highly destructive, as altering a single edge can inadvertently change the satisfiability status of the underlying formula. Consequently, augmentation for NP-hard problems requires structure-preserving techniques. Our framework bypasses destructive perturbations entirely by generating valid, benchmark-aligned formulas from the ground up, providing a domain-native augmentation strategy that scales combinatorial learning safely.

We discuss additional related work in \Cref{app:related_work}.

\section{Preliminaries}
\label{sec:preliminaries}

% \paragraph{Boolean Satisfiability and CNF.}
% Boolean satisfiability (SAT) is the decision problem of determining whether a Boolean formula admits an assignment of truth values that makes the formula true. Throughout this paper we consider formulas in \emph{conjunctive normal form} (CNF), in which a formula is written as a conjunction of clauses and each clause is a disjunction of literals. A literal is either a Boolean variable $p_j$ or its negation $\neg p_j$. Thus a CNF formula has the form
% \[
% \varphi = \bigwedge_{i=1}^{m} C_i, \qquad C_i = \bigvee_{\ell \in \mathcal{L}_i} \ell.
% \]
% A formula is satisfiable (SAT) if there exists an assignment under which every clause evaluates to true; otherwise it is unsatisfiable (UNSAT). To illustrate the notations, the formula $(p_1 \vee \neg p_2) \wedge (p_2 \vee p_3)$ is satisfiable; for instance, the assignment $p_1=p_2=\mathrm{true}$ makes both clauses true. In contrast, the formula $(p_1) \wedge (\neg p_1)$ is unsatisfiable, since no assignment can satisfy both unit clauses simultaneously.

\paragraph{Boolean Satisfiability and CNF.} 
Boolean satisfiability (SAT) is the decision problem of determining whether a Boolean formula admits an assignment of truth values that makes the formula true. Throughout this paper, we consider formulas in \emph{conjunctive normal form} (CNF), where a formula $\varphi$ is a conjunction of $m$ clauses over $n$ binary variables $x_j \in \{0, 1\}$. Each clause $C_i$ is a disjunction of these variables or their negations. Thus:
\begin{equation}
\varphi = \bigwedge_{i=1}^{m} C_i.
\end{equation}

A formula is satisfiable (SAT) if there exists an assignment $\mathbf{x} \in \{0,1\}^n$ under which every clause evaluates to true. For example, $(x_1 \vee \neg x_2) \wedge (x_2 \vee x_3)$ is SAT (e.g., $x_1=1, x_2=1$), while $(x_1) \wedge (\neg x_1)$ is UNSAT.

% \paragraph{Binary Linear Encoding.}
% A CNF formula admits a standard binary linear encoding. By associating each Boolean variable $p_j$ with a binary decision variable $x_j \in \{0,1\}$, a clause with positive literals $P$ and negative literals $N$ is satisfied if and only if:
% \[
% \sum_{j \in P} x_j - \sum_{j \in N} x_j \ge 1 - N.
% \]
% For example, under this encoding, a clause such as $p_1 \vee \neg p_2 \vee p_3$ can be written as
% \[
% x_1 + (1-x_2) + x_3 \ge 1,
% \]
% and, after rearrangement, as
% \[
% x_1 - x_2 + x_3 \ge 0.
% \]
% Applying this conversion clause by clause maps a CNF formula $\varphi = \bigwedge_{i=1}^m C_i$ to an integer system of inequalities $A\mathbf{x} \ge \mathbf{b}$, with $A \in \{-1,0,1\}^{m \times n}$ and $\mathbf{b} \in \mathbb{Z}^m$. Satisfiability is therefore exactly equivalent to binary feasibility: the formula is SAT if and only if there exists $\mathbf{x} \in \{0,1\}^n$ satisfying all inequalities.

\paragraph{Binary Linear Encoding.}
A CNF formula admits a standard binary linear encoding. By associating each Boolean variable with a binary decision variable, a clause with positive variables $P$ and negative variables $N$ is satisfied if and only if
\begin{equation}
    \sum_{j \in P} x_j - \sum_{j \in N} x_j \ge 1 - N.
\end{equation}

For example, under this encoding, a clause such as $x_1 \vee \neg x_2 \vee x_3$ can be written as
\begin{equation}
x_1 + (1-x_2) + x_3 \ge 1,
\end{equation}
and, after rearrangement, as
\begin{equation}
x_1 - x_2 + x_3 \ge 0.
\end{equation}

Applying this conversion clause by clause maps a CNF formula $\varphi = \bigwedge_{i=1}^m C_i$ to an integer system of inequalities $A\mathbf{x} \ge \mathbf{b}$, with $A \in \{-1,0,1\}^{m \times n}$ and $\mathbf{b} \in \mathbb{Z}^m$. Satisfiability is therefore exactly equivalent to binary feasibility: the formula is SAT if and only if there exists $\mathbf{x} \in \{0,1\}^n$ satisfying all inequalities.

% \paragraph{Slack Variables and Clause Semantics.}
% For both analysis and model design, it is convenient to rewrite the inequality system in equality form by introducing a nonnegative integer slack vector $\mathbf{s} \in \mathbb{Z}_+^m$, yielding:
% \[
% A\mathbf{x} - \mathbf{s} = \mathbf{b}.
% \]
% Crucially, the slack variable $s_i$ is not merely an algebraic artifact; it has a direct semantic interpretation. It records the number of \emph{excess} literals satisfying clause $C_i$ beyond the single literal required for feasibility. Consequently, $s_i = 0$ exactly when the clause is satisfied by a single literal, whereas $s_i > 0$ when it is satisfied by two or more literals.

\paragraph{Slack Variables and Clause Semantics.}
For both analysis and model design, it is convenient to rewrite the inequality system in equality form by introducing a slack vector $\mathbf{s}$ with $s_i \in \{0, \dots, k_i - 1\}$ for each clause $C_i$ of width $k_i$, yielding:
\begin{equation}
A\mathbf{x} - \mathbf{s} = \mathbf{b}.
\end{equation}
The slack variable $s_i$ is not merely an algebraic artifact; it has a direct semantic interpretation. It records the number of literals satisfying clause $C_i$ beyond the single literal required for feasibility. Consequently, $s_i = 0$ exactly when the clause is satisfied by a single literal, whereas $s_i > 0$ when it is satisfied by two or more literals.

\section{Target-Aware Solver-Free Data Generation}\label{sec:datagen}

The data-generation pipeline is the primary practical component of our framework. Its starting point is the \emph{solution-first} viewpoint introduced by BPGNN \citep{eliasof2024}: rather than solving each instance to obtain a label, one first constructs an instance around a known solution or contradiction. However, this idea must respect \emph{CNF semantics}. Once the variables and polarities of a clause are fixed, its induced LP inequality is fixed as well; in particular, the LP right-hand side is not a free continuous parameter that can be chosen independently. For this reason, our generator is \emph{CNF-native}: we generate clauses directly in formula space and only afterwards map the resulting CNF into an integer linear program (ILP). This design preserves logical validity at the data-creation stage while also matching the representation consumed by the downstream GNN.

% Both the SAT and UNSAT generators are fully \emph{residual-aligned}: clause polarities are chosen deterministically to match the LP slack distribution of the downstream benchmark, so the residual signals seen during pretraining are already calibrated to those seen at test time.
% The generator is \emph{solver-free} (labels certified by construction), \emph{CNF-native} (output is a valid CNF formula), and \emph{algebraically consistent} (the same clause-to-ILP mapping is used for both generated and benchmark instances).

\begin{algorithm}[t]
\caption{SAT Generation}
\label{alg:target-aware-sat}
\begin{algorithmic}[1]
\REQUIRE Size $n$, clauses $m$, target benchmarks statistics $\mathcal{T}$, SAT slack dist.\ $\Pi_s$
% \ENSURE CNF formula $\varphi$, planted witness $\mathbf{x}^\star$, induced slack $\mathbf{s}^\star$, label \textsc{SAT}
\STATE Sample $\mathbf{x}^\star \in \{0,1\}^n$ uniformly at random
\FOR{$i = 1$ to $m$}
    \STATE Sample width $k_i$ from width distribution in $\mathcal{T}$
    \STATE Sample $k_i$ distinct variables, weighted by occurrence skew in $\mathcal{T}$
    \STATE Sample $s_i^\star \in \{0,\dots,k_i{-}1\}$ from $\Pi_s$ 
    \STATE Assign polarities to the first $(s_i^\star{+}1)$ selected variables so their literals evaluate to True under $\mathbf{x}^\star$; set polarities of the rest to evaluate to False
    \STATE Add resulting clause $C_i$ to $\varphi$
\ENDFOR
\RETURN $(\varphi,\, \mathbf{x}^\star,\, \mathbf{s}^\star,\, \textsc{SAT})$
\end{algorithmic}
\end{algorithm}

\subsection{SAT Instance Generation}\label{sec:sat-gen}

The SAT generation process is designed to match the induced LP slack distribution of a given benchmark. Once $\mathbf{x}^\star$ and a clause $C_i$ are fixed, the induced slack is determined:
\begin{equation}\label{eq:induced-slack}
    s_i^\star = A_i \mathbf{x}^\star - b_i = \#\{\text{literals in } C_i \text{ satisfied by } \mathbf{x}^\star\} - 1.
\end{equation}

The slack-planted generator eliminates any remaining mismatch by deterministically constructing each clause to match a target slack $s_i^\star$ drawn from the empirical benchmark SAT slack distribution. For a clause of width $k_i$ with target slack $s_i^\star$, we sample $k_i$ distinct variables weighted by the target benchmark's occurrence skew, and shuffle them. We then designate the first $(s_i^\star + 1)$ variables to be \emph{satisfied} (setting their polarity to match $\mathbf{x}^\star$) and the remaining $(k_i - s_i^\star - 1)$ variables to be \emph{violated} (setting their polarity opposite to $\mathbf{x}^\star$).

This generation process avoids the costly rejection loops inherent to solver-in-the-loop pipelines. Since the assignments are applied independently to each of the $m$ clauses up to width $k$, the computational complexity of producing a perfectly aligned, guaranteed-SAT instance is strictly $\mathcal{O}(mk)$. \Cref{alg:target-aware-sat} gives the complete procedure.

\begin{remark}[Non-uniqueness of satisfying assignments]\label{rem:multiple-solutions}
A planted witness does not imply a unique satisfying assignment. For SAT instances, existence of \emph{some} satisfying assignment is sufficient. The witness is used only to generate a correctly labeled formula and to characterize the induced slack assignment presented in \eqref{eq:induced-slack}.
\end{remark}

\subsection{UNSAT Instance Generation}\label{sec:unsat_generation}

For UNSAT instances, there is no globally satisfying planted witness. Instead, we certify unsatisfiability by construction. Each instance embeds a small UNSAT \emph{core} within a larger set of slack-planted \emph{filler} clauses. The core guarantees the \textsc{UNSAT} label; adding arbitrary clauses to an already-UNSAT formula preserves unsatisfiability.

To guarantee a contradiction, we select a dominant clause width $w$ and pick $w$ fresh variables $x_1, \dots, x_w$. We then construct an UNSAT core by generating all $2^w$ possible polarity combinations of these variables as clauses which is UNSAT by construction. In this case, the planted reference assignment $\mathbf{x}^\star$ will violate exactly one clause in the core.

\paragraph{Slack-planted filler.}
Just as the SAT generator plants the slack distribution $\mathbf{s}^\star$, the UNSAT filler generator aligns with the benchmark UNSAT slack distribution by sampling  
% Since UNSAT instances have no true global solution, we sample $\mathbf{x}^\star \in \{0,1\}^n$ uniformly to serve as a planted reference assignment, not a solution (as it violates the core), but an anchor for polarity construction.
a target slack $s_i^\star$ from the benchmark UNSAT slack distribution. 
% The motivation is the same as for the SAT side: during training on UNSAT benchmark instances, the model's internal residual signals converge toward the benchmark's apparent-slack distribution. 
% Pretraining on data with a matched apparent-slack distribution reduces the residual-level domain shift.
Our method is presented in \Cref{alg:ta-unsat}, and we give a simple example of \Cref{alg:target-aware-sat,alg:ta-unsat} in \Cref{app:set_unsat_example}

\begin{algorithm}[t]
\caption{UNSAT Generation}
\label{alg:ta-unsat}
\begin{algorithmic}[1]
\REQUIRE Size $n$, clauses $m$, target benchmark statistics $\mathcal{T}$, UNSAT slack dist.\ $\Pi_u$
\STATE Sample $\mathbf{x}^\star \in \{0,1\}^n$ uniformly at random
\STATE $w \gets$ dominant width in $\mathcal{T}$;\;\; pick $w$ fresh variables $x_1, \dots, x_w$
\STATE \textbf{Core construction:} Add $2^w$ clauses to $\psi$ comprising every possible combination of polarities for $x_1, \dots, x_w$ \hfill\textit{\small // Creates an UNSAT core; $\mathbf{x}^\star$ violates exactly 1 clause}
\FOR{$i = 1$ to $m - 2^w$}
    \STATE Sample width $k_i$ from width distribution in $\mathcal{T}$
    \STATE Sample $k_i$ distinct variables, weighted by occurrence skew in $\mathcal{T}$
    \STATE Sample $s_i^\star \in \{0,\dots,k_i{-}1\}$ from $\Pi_u$
    \STATE Assign polarities to the first $(s_i^\star{+}1)$ selected variables so their literals evaluate to True under $\mathbf{x}^\star$; set polarities of the rest to evaluate to False 
    \STATE Add resulting clause $C_i$ to $\psi$
\ENDFOR
\RETURN $(\psi,\, \mathbf{x}^\star,\, \mathbf{s}^\star,\, \textsc{UNSAT})$
\end{algorithmic}
\end{algorithm}

% \paragraph{Relation to $\mathbf{x}$ and $\mathbf{s}$.}
% SAT generation is \emph{witness-first}: $\mathbf{x}^\star$ is planted and $\mathbf{s}^\star$ is induced. UNSAT generation is \emph{proof-first}: no binary $\mathbf{x}$ admits a nonneg slack vector satisfying $A\mathbf{x} - \mathbf{s} = \mathbf{b}$. The dummy $\mathbf{x}_{\mathrm{ref}}$ in the slack-planted UNSAT generator is used only to control the polarity distribution of filler clauses; it is not a solution and is not recorded as part of the generated instance.

\section{Linear-Programming Graph Neural Network (LPGNN)}

Given a CNF formula $\varphi$ with $n$ Boolean variables and $m$ clauses, we first map it to the standard binary linear feasibility problem
\begin{equation}
    A\mathbf{x} \geq \mathbf{b},
    \qquad
    \mathbf{x}\in\{0,1\}^n,
    \label{eq:cnf-ilp}
\end{equation}
where each row of $A\in\{-1,0,1\}^{m\times n}$ corresponds to one clause. We then introduce nonnegative slack variables and write the system in equality form
\begin{equation}
    \widehat A \mathbf{z} = \mathbf{b},
    \qquad
    \widehat A = [\,A\mid -I\,],
    \qquad
    \mathbf{z}
    =
    \begin{bmatrix}
        \mathbf{x}\\
        \mathbf{s}
    \end{bmatrix}
    .
    \label{eq:augmented-system}
\end{equation}
% Strictly speaking, since the original inequalities are of the form $A\mathbf{x}\geq \mathbf{b}$, the variables $\mathbf{s}$ are surplus variables. We use the term \emph{slack} throughout for simplicity.

The LPGNN is built on the augmented variable vector $\mathbf{z}=[\mathbf{x};\mathbf{s}]$. Thus, the graph contains two node types:
\begin{equation}
    V = V_x \cup V_s,
    \qquad
    |V_x|=n,
    \qquad
    |V_s|=m.
\end{equation}
The node $v_j\in V_x$ represents Boolean variable $x_j$, and the node $u_i\in V_s$ represents the slack variable $s_i$ associated with clause $i$. We include an edge $(u_i,v_j)$ whenever $A_{ij}\neq 0$, with edge attribute $A_{ij}$. 
% Optionally, we also include variable--variable edges between variables that co-occur in a clause. 
% The bipartite variable--slack graph is the default backbone; the variable-interaction edges are treated as an architectural augmentation.

Each node is initialized from simple structural features. For variable nodes, these include degree, positive occurrence count, and negative occurrence count. For slack nodes, these include clause width, right-hand side $b_i$, and literal-composition statistics. Let
    $\mathbf{q}_p \in \mathbb{R}^{d_{\rm in}}$
denote the raw feature vector of node $p\in V$. The initial hidden state is
\begin{equation}
    \mathbf{h}^{(0)}_p
    =
    \psi_{\rm init}(\mathbf{q}_p)
    \in \mathbb{R}^d,
    \qquad p=1,\ldots,n+m,
    \label{eq:init}
\end{equation}
where $\psi_{\rm init}$ is an MLP shared across nodes, possibly with node-type embeddings.

\paragraph{Layer update.}
At layer $\ell$, the hidden states are collected in
\begin{equation}
    H^{(\ell)}
    =
    \begin{bmatrix}
        (\mathbf{h}_1^{(\ell)})^\top\\
        \vdots\\
        (\mathbf{h}_{n+m}^{(\ell)})^\top
    \end{bmatrix}
    \in \mathbb{R}^{(n+m)\times d}.
\end{equation}
The layer has two components: a standard graph message-passing update and an LP residual update.

First, each node embedding is projected to a scalar continuous surrogate for the corresponding coordinate of $\mathbf{z}$:
\begin{equation}
    \tilde z_p^{(\ell)}
    =
    \sigma\!\left(
        \alpha^{(\ell)}(\mathbf{h}_p^{(\ell)})
    \right),
    \qquad
    p=1,\ldots,n+m,
    \label{eq:latent-z}
\end{equation}
where $\alpha^{(\ell)}:\mathbb{R}^d\to\mathbb{R}$ is a learned scalar projection and $\sigma$ is the activation function. In vector form,
\begin{equation}
    \tilde{\mathbf z}^{(\ell)}
    =
    \sigma\!\left(\alpha^{(\ell)}(H^{(\ell)})\right)
    \in \mathbb{R}^{n+m}.
\end{equation}
The vector $\tilde{\mathbf z}^{(\ell)}$ is not interpreted as a valid discrete assignment. It is a continuous latent surrogate used only to compute feasibility-related features.

The clause-level residual is
\begin{equation}
    \mathbf{r}^{(\ell)}
    =
    \widehat A\tilde{\mathbf z}^{(\ell)}-\mathbf{b}
    \in \mathbb{R}^m.
    \label{eq:layer-residual}
\end{equation}
This residual lives in constraint space. Since the GNN updates node embeddings associated with the coordinates of $\mathbf{z}$, we map the residual back to node space using the transpose:
\begin{equation}
    \mathbf{g}^{(\ell)}
    =
    \widehat A^\top \mathbf{r}^{(\ell)}
    \in \mathbb{R}^{n+m}.
    \label{eq:node-residual}
\end{equation}
The $p$th entry $g_p^{(\ell)}$ aggregates the signed residuals of all constraints in which node $p$ participates. Equivalently, if one temporarily relaxes $\mathbf{z}$ to a continuous variable, then
\begin{equation}
    \widehat A^\top \mathbf{r}^{(\ell)}
    =
    \nabla_{\mathbf z}
    \frac{1}{2}
    \|\widehat A\mathbf z-\mathbf b\|_2^2
    \bigg|_{\mathbf z=\tilde{\mathbf z}^{(\ell)}}.
\end{equation}
We emphasize that this is not a gradient of the original discrete SAT problem. It is a continuous feasibility signal projected from constraint space back to node space.

The scalar node residual is lifted to the hidden dimension by another learned map:
\begin{equation}
    \mathbf{e}_p^{(\ell)}
    =
    \rho^{(\ell)}\!\left(g_p^{(\ell)}\right)
    \in \mathbb{R}^d,
    \qquad
    \rho^{(\ell)}:\mathbb{R}\to\mathbb{R}^d.
    \label{eq:residual-lift}
\end{equation}
We then combine this LP-aware signal with standard message passing. Let $\mathcal{N}(p)$ denote the neighbors of node $p$, and let $a_{pq}$ denote the edge attribute, e.g. $A_{ij}$ for variable--slack edges. A generic LPGNN layer is
% \begin{align}
%     \mathbf{m}_p^{(\ell)}
%     &=
%     \sum_{q\in\mathcal{N}(p)}
%     \phi_{\rm msg}^{(\ell)}
%     \left(
%         \mathbf{h}_p^{(\ell)},
%         \mathbf{h}_q^{(\ell)},
%         a_{pq}
%     \right),
%     \label{eq:message}
%     \\
%     \bar{\mathbf h}_p^{(\ell)}
%     &=
%     \mathbf{h}_p^{(\ell)}
%     +
%     \mathbf{e}_p^{(\ell)},
%     \label{eq:residual-injection}
%     \\
%     \mathbf{h}_p^{(\ell+1)}
%     &=
%     \phi_{\rm upd}^{(\ell)}
%     \left(
%         \bar{\mathbf h}_p^{(\ell)},
%         \mathbf{m}_p^{(\ell)}
%     \right).
%     \label{eq:update}
% \end{align}

\begin{equation}
\label{eq:update}
\begin{aligned}
    \mathbf{m}_p^{(\ell)}
    &=
    \sum_{q\in\mathcal{N}(p)}
    \phi_{\rm msg}^{(\ell)}
    \left(
        \mathbf{h}_p^{(\ell)},
        \mathbf{h}_q^{(\ell)},
        a_{pq}
    \right),
    \\
    \bar{\mathbf h}_p^{(\ell)}
    &=
    \mathbf{h}_p^{(\ell)}
    +
    \mathbf{e}_p^{(\ell)},
    \qquad
    \mathbf{h}_p^{(\ell+1)}
    =
    \phi_{\rm upd}^{(\ell)}
    \left(
        \bar{\mathbf h}_p^{(\ell)},
        \mathbf{m}_p^{(\ell)}
    \right).
\end{aligned}
\end{equation}

Here $\phi_{\rm msg}^{(\ell)}$ and $\phi_{\rm upd}^{(\ell)}$ are learned neural networks. In practice, $\phi_{\rm upd}^{(\ell)}$ may be implemented as an MLP-style update, residual block, or attention-based update. The essential feature of LPGNN is not the specific message-passing primitive, but the additional node-space feasibility signal $\widehat A^\top \mathbf r^{(\ell)}$ injected at every layer.

Equivalently, in matrix notation, one may write a layer abstractly as
\begin{equation}
    H^{(\ell+1)}
    =
    \mathrm{MPNN}^{(\ell)}
    \left(
        H^{(\ell)}
        +
        \rho^{(\ell)}
        \left(
            \widehat A^\top
            \left[
                \widehat A
                \sigma\!\left(\alpha^{(\ell)}(H^{(\ell)})\right)
                -
                \mathbf b
            \right]
        \right),
        G
    \right),
    \label{eq:compact-layer}
\end{equation}
where $\rho^{(\ell)}$ is applied row-wise to the scalar node residuals, and MPNN stands for the message passing neural network framework.

\paragraph{Slack nodes versus constraint nodes.}
In a standard bipartite SAT graph implementation, constraint nodes act as summaries of clause. In contrast, in LPGNN the slack node $s_i$ is not a representation of clause $i$, but an explicit component of the augmented variable vector $\mathbf{z} = [\mathbf{x}; \mathbf{s}]$. As a result, slack nodes participate directly in the residual computation~\eqref{eq:layer-residual} and receive feasibility signals in node space via~\eqref{eq:node-residual}.
This design induces a tight coupling between message passing and constraint satisfaction. The architecture updates the latent representations of both variables and slacks, which together determine the constraint residuals.

\section{Experiments}
\label{sec:experiments}

\subsection{Generator Efficiency}\label{sec:gen-efficiency}

\begin{table}[t!]
\centering
% \caption{Median wall-clock time (ms) to produce one correctly labeled SAT/UNSAT pair. $n$ = variables, $m = \lfloor 4.258\,n \rceil$ = clauses (phase-transition ratio; nearest integer). \textbf{Naive} = brute-force $2^n$ assignment enumeration (theoretical lower bound; not a deployed method); \textbf{CaDiCaL} = random 3-CNF sampled until SAT, then until UNSAT, using CaDiCaL for each satisfiability check (labeling only; excludes DRAT-trim proof verification used in G4SATBench for UNSAT instances---including that cost would increase the baseline time and the speedup); \textbf{Ours} = target-aware planted-solution generation (no solver; benchmark statistics extracted once as a preprocessing step, not included in reported times). Speedup = CaDiCaL / Ours. 200 repetitions, median; Naive capped at 20 reps / 30\,s timeout (N/A = infeasible beyond $n{=}20$). At $n{\geq}500$, CaDiCaL did not complete within a practical budget; \textsuperscript{†}values extrapolated via log-linear fit to the $n{=}100$--$250$ measured data (doubling every ${\approx}17$ variables), giving ${\sim}10^8$\,ms at $n{=}500$ and ${\sim}10^{16}$\,ms at $n{=}1000$.}
\caption{Wall-clock time (ms) to generate one correctly labeled SAT/UNSAT pair at the phase transition. \textbf{Naive} enumerates all $2^n$ assignments (capped at a 30s timeout). \textbf{CaDiCaL} represents the standard solver-in-the-loop pipeline (generate-and-test). \textbf{Ours} evaluates the proposed solver-free, target-aware generation. We present median over 200 repetitions. For $n \ge 500$, CaDiCaL times are extrapolated via log-linear fit from measured data.}
\label{tab:gen-efficiency}
\begin{tabular}{rrcccr}
\toprule
$n$ & $m$ & Naive (ms) & CaDiCaL (ms) & Ours (ms) & Speedup \\
\midrule
  15  &    64  & 272      & 0.66  & 0.54 & $1.2\times$ \\
  20  &    85  & 6{,}188  & 0.87  & 0.71 & $1.2\times$ \\
\midrule
  50  &   213  & N/A & 2.26   & 1.81 & $1.2\times$ \\
  75  &   319  & N/A & 4.47   & 2.71 & $1.6\times$ \\
 100  &   426  & N/A & 8.98   & 3.64 & $2.5\times$ \\
 150  &   639  & N/A & 66.1   & 5.47 & $12\times$ \\
 200  &   852  & N/A & 525    & 7.42 & $71\times$ \\
 250  &  1064  & N/A & 4{,}080 & 9.32 & $438\times$ \\
 500  &  2129  & N/A & ${\sim}10^{8}$\textsuperscript{†} & 19.3 & ${\sim}10^{6}\times$ \\
1000  &  4258  & N/A & ${\sim}10^{16}$\textsuperscript{†} & 39.0 & ${\sim}10^{14}\times$ \\
\bottomrule
\end{tabular}
\end{table}

% \begin{figure}[t]
%     \centering
% \includegraphics[width=0.8\textwidth]{figures/generation_cost_vs_scale.pdf}
%     \caption{Generation cost.}
%     \label{fig:gen_cost}
% \end{figure}

% \begin{figure}[t]
%     \centering
% \includegraphics[width=0.8\textwidth]{figures/slack_distribution_comparison.pdf}
%     \caption{Generation data follows data slack distribution (full instances for sat, non-core instances for unsat).}
%     \label{fig:gen_cost}
% \end{figure}

\begin{figure}[h]
    \centering
    \includegraphics[width=0.8\textwidth]{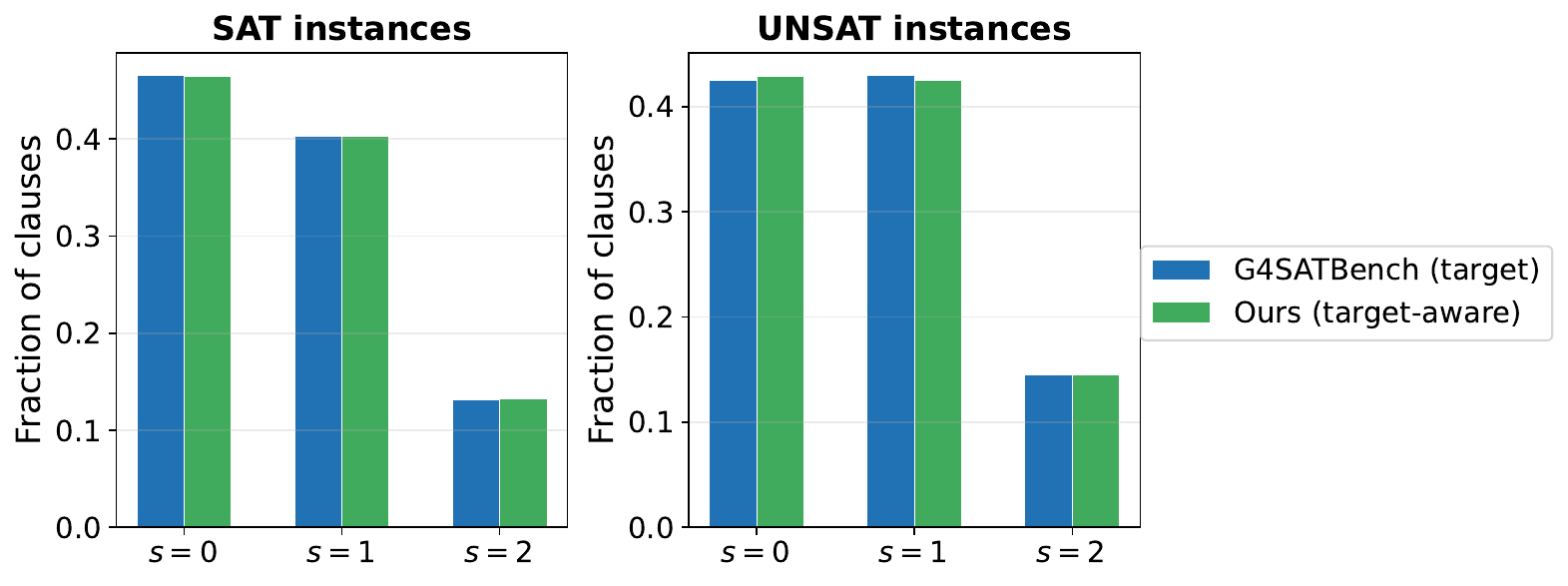}
    \caption{Comparison of the induced slack distributions between the target G4SATBench dataset and our synthetic instances. The generated data faithfully mirrors the target statistics (measured on full instances for SAT, and non core-contradiction filler clauses for UNSAT).}
    \label{fig:slack_dist_comparison}
\end{figure}

\begin{figure}[ht]
    \centering
    
    \begin{subfigure}[b]{0.48\textwidth}
        \centering
        \includegraphics[width=\textwidth]{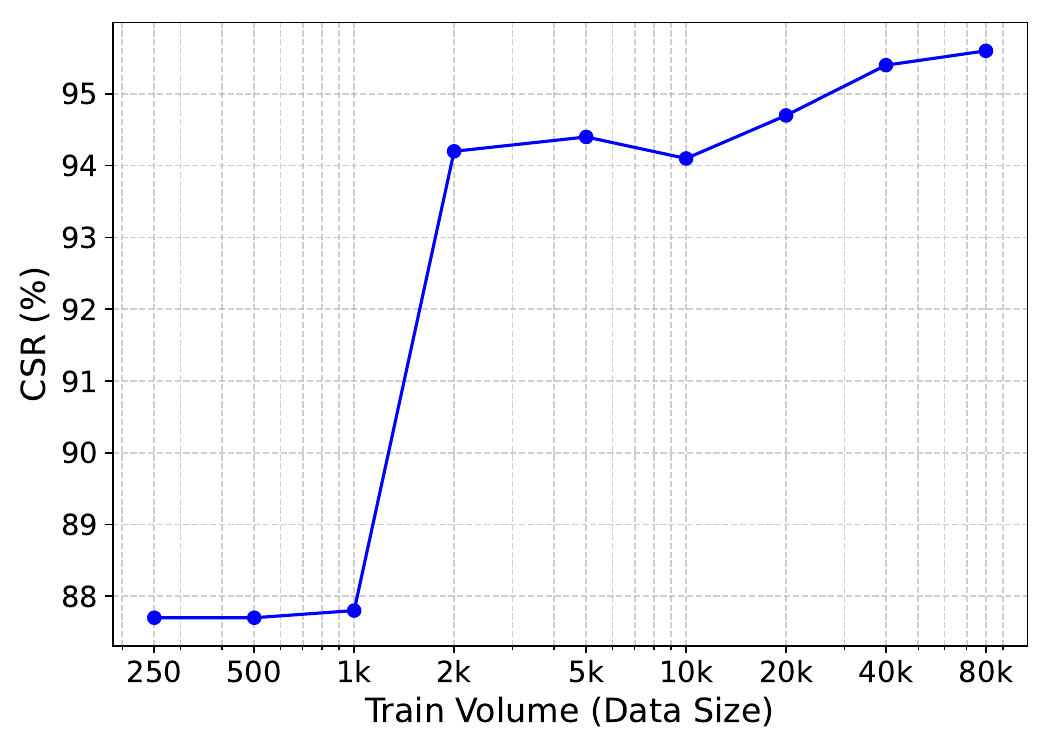}
        \caption{CSR.}
        \label{fig:csr_16l_medium}
    \end{subfigure}
    \hfill
    \begin{subfigure}[b]{0.48\textwidth}
        \centering
        \includegraphics[width=\textwidth]{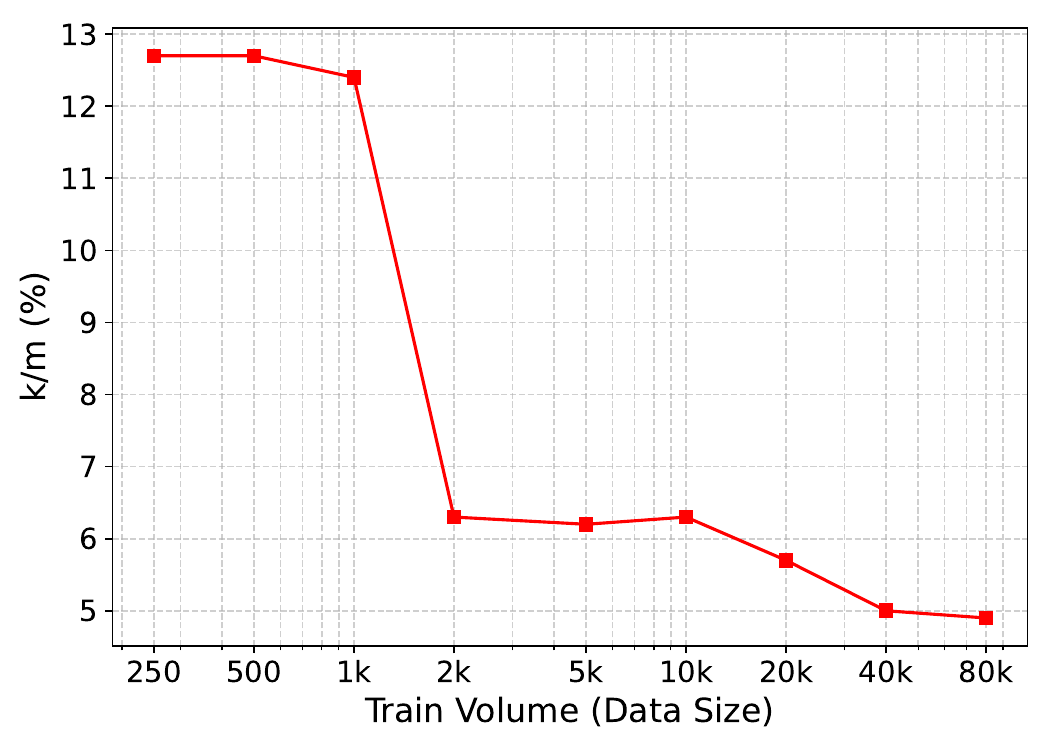}
        \caption{$k/m$ ratio.}
        \label{fig:km_16l_medium}
    \end{subfigure}
    
    \caption{Scaling behavior of the LPGNN trained on generated data as a function of training data volume. Evaluated on the G4SATBench medium 3-SAT test set. As the volume of training data increases, the Constraint Satisfaction Rate (CSR) consistently improves, indicating higher-quality assignment predictions for SAT instances. Concurrently, the $k/m$ ratio decreases, demonstrating that the predicted assignments for UNSAT instances violate progressively fewer clauses.}
    \label{fig:scaling_16l_medium_combined}
\end{figure}

We evaluate the computational efficiency of our solver-free generation method against a standard solver-in-the-loop pipeline, and show the results in \Cref{tab:gen-efficiency}. The \emph{naive} baseline (brute-force enumeration) illustrates the theoretical worst-case exponential scaling, becoming computationally infeasible beyond small formula sizes. The CaDiCaL \citep{biere2020cadical} baseline represents a conventional dataset construction approach: repeatedly sampling random 3-CNF formulas near the satisfiability phase transition, that is, the critical clause-to-variable ratio at which instances sharply shift from being mostly satisfiable to mostly unsatisfiable \citep{crawford1996experimental}, and applying the CaDiCaL solver to determine satisfiability.

Our proposed target-aware generator scales near-linearly with the number of clauses, as satisfiability or unsatisfiability is guaranteed by construction without solver verification. While initialization and slack-matching overhead make generation times comparable to CaDiCaL for small instances, the proposed method demonstrates significant advantages at benchmark-relevant scales. At large scales, the solver-in-the-loop approach becomes infeasible, while our solver-free mechanism shows scaling up stability.
% These reported speedups represent a conservative lower bound. The solver baseline omits proof verification costs for UNSAT instances and ignores the increasingly prohibitive rejection rates of random sampling at larger $n$. 
% Furthermore, these reported speedups are conservative. The solver baseline excludes the cost of proof generation and verification for UNSAT instances, and does not account for the sharply increasing rejection rates of random sampling near the phase transition as $n$ grows. Incorporating these effects would further widen the efficiency gap in favor of our approach.
Furthermore, these reported speedups are conservative, as the solver baseline excludes the cost of proof generation and verification for UNSAT instances. Incorporating this effect would further widen the efficiency gap in favor of our approach. \Cref{app:generation_cost_plot} provides a visual representation of the scaling results presented in \Cref{tab:gen-efficiency}.

\subsection{Scaling Behavior}
\label{subsec:experiments_scaling_ood}

We utilize the G4SATBench dataset \citep{li2024g4satbench} for our experiments. We first verify that the generated 3-SAT instances reproduce the empirical slack distribution of the target benchmark (Figure~\ref{fig:slack_dist_comparison}). To evaluate the scalability and assignment quality of the proposed architecture, we train the LPGNN model on these generated 3-SAT instances. We investigate the impact of training data volume on the model's capacity to predict assignments. Performance is quantified via the Constraint Satisfaction Rate (CSR) of SAT instances and the violated-clause fraction ($k/m$) of UNSAT instances, where $k$ is the number of violated clauses in the predicted assignment, and $m$ represents the total number of clauses in the formula.

We evaluate the model's performance on the in-distribution G4SATBench test sets. The scaling trajectories for models trained and evaluated on the 3-SAT distributions are illustrated in  \Cref{fig:scaling_16l_medium_combined}. 
The empirical results demonstrate an improvement in assignment quality as a function of training volume. 
% Specifically, on the medium distribution (Figure~\ref{fig:scaling_16l_combined}, top row), scaling the dataset from 250 to 80,000 instances increases the CSR from 87.7\% to 95.6\%. Simultaneously, the $k/m$ ratio decreases from 12.7\% to 4.9\%. Comparable trends are observed for the easy distribution (Figure~\ref{fig:scaling_16l_combined}, bottom row). 
These pronounced scaling trajectories indicate that the LPGNN architecture effectively leverages expanded datasets to acquire robust assignments heuristics.

\begin{figure}[t]
    \centering
    % --- Left Side: Table ---
    \begin{minipage}[c]{0.48\textwidth}
        \centering
        % \captionof{table}{Scaling results for architectures trained on target-aware synthetic data, evaluated on the Medium 3-SAT test set. NL denotes \textbf{NLocalSAT} and QS denotes \textbf{QuerySAT}.}
        \captionof{table}{Scaling results for architectures trained on target-aware synthetic data, evaluated on the Medium 3-SAT test set. Both models exhibit improved assignment quality as data increases, reflected by a higher CSR and a lower $k/m$ ratio. NL denotes \textbf{NLocalSAT} and QS denotes \textbf{QuerySAT}.}
        \label{tab:baseline_scaling_medium}
        \small
        \begin{tabular}{c | cc | cc}
        \toprule
        \textbf{Train} & \multicolumn{4}{c}{\textbf{Medium 3-SAT}} \\
        \textbf{Volume} & \multicolumn{2}{c}{CSR (\%) $\uparrow$} & \multicolumn{2}{c}{$k/m$ (\%) $\downarrow$} \\
         & NL & QS & NL & QS \\
        \midrule
        250  & 94.4 & 91.8 & 6.2 & 8.8 \\
        500  & 94.0 & 93.9 & 6.8 & 6.9 \\
        1k   & 95.4 & 94.3 & 5.3 & 6.1 \\
        2k   & 98.2 & 94.5 & 2.4 & 6.1 \\
        5k   & 98.2 & 94.4 & 2.4 & 6.3 \\
        10k  & 99.0 & 94.4 & 1.6 & 6.4 \\
        \bottomrule
        \end{tabular}
    \end{minipage}
    \hfill
    % --- Right Side: Figure ---
    \begin{minipage}[c]{0.48\textwidth}
        \centering
        \includegraphics[width=\linewidth]{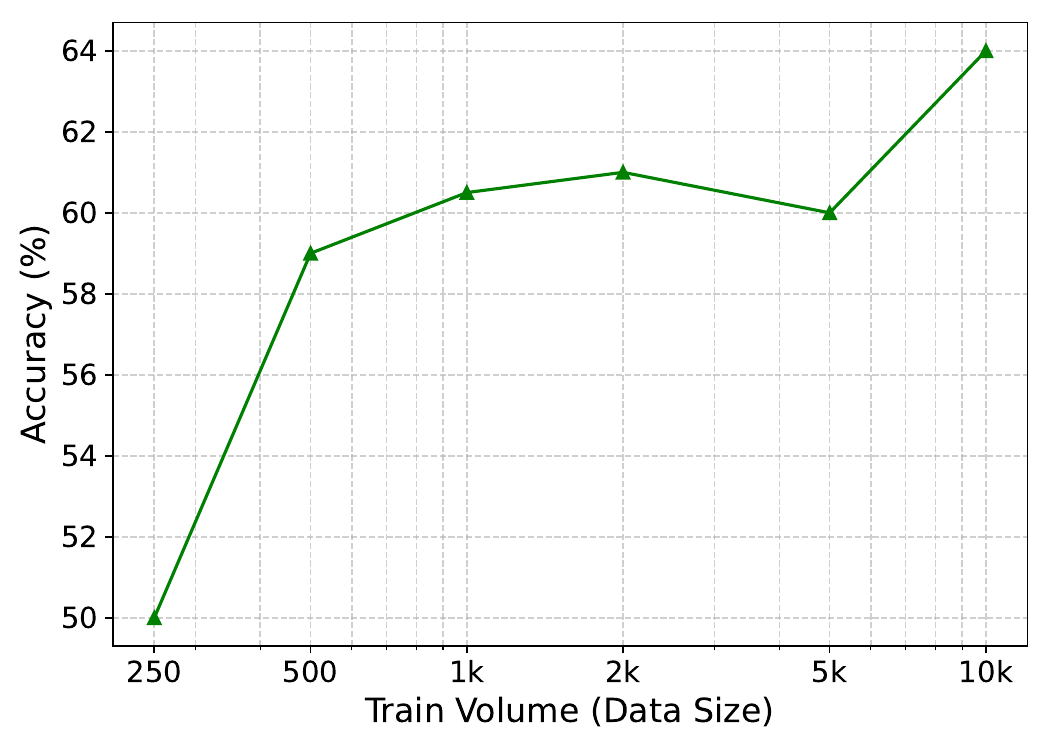}
        \captionof{figure}{Accuracy scaling of the LPGNN model evaluated on the Easy 3-SAT test set, for training volumes up to 10k instances.}
        \label{fig:ba_scaling_easy}
    \end{minipage}
\end{figure}

Importantly, the proposed data-generation framework is not restricted to a specific model; it is compatible with any architecture equipped with an assignment prediction head. To validate this claim, we provide results in \Cref{tab:baseline_scaling_medium} for two alternative architectures: NLocalSAT \citep{zhang2020nlocalsat} and QuerySAT \citep{ozolins2022}. These results demonstrate that our benchmark-aligned synthetic data generalizes across different models, improving performance across various architectures.

In addition to assignment quality, we examine the model's classification performance as a function of dataset size. As illustrated in \Cref{fig:ba_scaling_easy}, classification accuracy improves with training volume, evolving from near-random guessing at the smallest data scales to a stronger predictive signal as more data is introduced.

% To further evaluate our framework, we provide additional empirical results in \Cref{app:additional_results}. These include generalization tests on other benchmark instances, scaling comparisons against generic data, and mixed-training experiments with the original benchmark. We show ablation study isolating the impact of the LP residuals in \Cref{app:ablation_study}.
% We provide additional dataset details in \Cref{app:dataset_statistics} and further experimental settings in \Cref{app:experiments_detail}. 
% Finally, we discuss the limitations of our framework in \Cref{app:limitations} and its broader societal impacts in \Cref{app:broader_impacts}.

To further evaluate our framework, we provide additional empirical results in \Cref{app:additional_results}. These include generalization tests on other benchmark instances, scaling comparisons against generic data, and mixed-training experiments with the original benchmark. Additionally, we present an ablation study isolating the impact of the LP residuals in \Cref{app:ablation_study}. Finally, comprehensive dataset details and further experimental settings are provided in \Cref{app:dataset_statistics}.

\section{Conclusion}

Learning-based approaches to NP-hard problems are often bottlenecked not by model capacity, but by the cost of obtaining labeled data. We have introduced a target-aware, solver-free data generation framework that eliminates the reliance on expensive solver-in-the-loop labeling, enabling the efficient production of large-scale, correctly labeled SAT data. By constructing instances with certified labels and aligning them to benchmark structure, our approach provides both computational efficiency and task-relevant supervision. Empirically, this translates into orders-of-magnitude speedups in data generation and clear performance gains when training GNNs. We additionally discuss the scope of this framework and broader societal impacts in \Cref{app:limitations} and \Cref{app:broader_impacts}.

More broadly, our results support a data-centric perspective on learning NP-hard problems, highlighting that efficient, distribution-aware data generation is a key enabler for unlocking the potential of modern machine learning methods. Building on this foundation, future work will explore expanding this generation framework to capture the complex, hidden community structures required for specialized industrial SAT applications, as well as adapting similar target-aware methodology to synthesize training data for other NP-hard combinatorial optimization problems. Ultimately, this work demonstrates that scalable, task-aligned data generation is just as critical as model architecture in advancing deep learning for complex combinatorial reasoning.

\clearpage
\newpage

\bibliographystyle{plainnat}
\bibliography{references}

@techreport{salari2000code,
  title={Code verification by the method of manufactured solutions},
  author={Salari, Kambiz and Knupp, Patrick},
  institution={Sandia National Laboratories},
  number={SAND2000-1444},
  year={2000},
  doi={10.2172/759450}
}

@inproceedings{kingma2015adam,
  title={Adam: A Method for Stochastic Optimization},
  author={Kingma, Diederik P and Ba, Jimmy},
  booktitle={International Conference on Learning Representations (ICLR)},
  year={2015}
}

@article{crawford1996experimental,
  title={Experimental results on the crossover point in random 3-SAT},
  author={Crawford, James M and Auton, Larry D},
  journal={Artificial Intelligence},
  volume={81},
  number={1-2},
  pages={31--57},
  year={1996},
  publisher={Elsevier}
}

@inproceedings{biere2020cadical,
    title = "CaDiCaL, Kissat, Paracooba, Plingeling and Treengeling Entering the SAT Competition 2020",
    author = "Armin Biere and Katalin Fazekas and Mathias Fleury and Maximilian Heisinger",
    year = "2020",
    language = "English",
    volume = "vol. B-2020-1",
    series = "Department of Computer Science Report Series B",
    publisher = "University of Helsinki",
    pages = "50--53",
    booktitle = "Proc. of SAT Competition 2020 - Solver and Benchmark Descriptions",
}

@inproceedings{Moskewiczetal2001,
  author    = {Moskewicz, Matthew W. and Madigan, Conor F. and Zhao, Ying and Zhang, Lintao and Malik, Sharad},
  title     = {Chaff: Engineering an Efficient SAT Solver},
  booktitle = {Proceedings of the 38th Annual Design Automation Conference (DAC)},
  year      = {2001},
  pages     = {530--535},
  doi       = {10.1145/378239.379017}
}

@article{Soosetal2009,
  author    = {Soos, Mate and Nohl, Karsten and Castelluccia, Claude},
  title     = {Extending SAT Solvers to Cryptographic Problems},
  journal   = {Lecture Notes in Computer Science},
  year      = {2009},
  volume    = {5584},
  pages     = {244--257},
  doi       = {10.1007/978-3-642-02777-2_24}
}

@inproceedings{Biereetal1999,
  author    = {Biere, Armin and Cimatti, Alessandro and Clarke, Edmund and Zhu, Yunshan},
  title     = {Symbolic Model Checking without BDDs},
  booktitle = {Tools and Algorithms for the Construction and Analysis of Systems (TACAS)},
  year      = {1999},
  pages     = {193--207},
  doi       = {10.1007/3-540-49059-0_14}
}

@inproceedings{zhang2020nlocalsat,
  title={NLocalSAT: Boosting local search with solution prediction},
  author={Zhang, Wenjie and Sun, Zeyu and Zhu, Qihao and Li, Ge and Cai, Shaowei and Xiong, Yingfei and Zhang, Lu},
  journal={arXiv preprint arXiv:2001.09398},
  year={2020}
}

@inproceedings{xu2019how,
  title={How Powerful are Graph Neural Networks?},
  author={Xu, Keyulu and Hu, Weihua and Leskovec, Jure and Jegelka, Stefanie},
  booktitle={International Conference on Learning Representations (ICLR)},
  year={2019}
}

@inproceedings{you2019g2sat,
  title={G2SAT: Learning to Generate {SAT} Formulas},
  author={You, Jiaxuan and Wu, Haoze and Barrett, Clark and Ramanujan, Raghuram and Leskovec, Jure},
  booktitle={Advances in Neural Information Processing Systems (NeurIPS)},
  volume={32},
  year={2019}
}

@article{kaplan2020scaling,
  title={Scaling Laws for Neural Language Models},
  author={Kaplan, Jared and McCandlish, Sam and Henighan, Tom and Brown, Tom B. and Chess, Benjamin and Child, Rewon and Gray, Scott and Radford, Alec and Wu, Jeffrey and Amodei, Dario},
  journal={arXiv preprint arXiv:2001.08361},
  year={2020}
}

@article{hoffmann2022training,
  title={Training Compute-Optimal Large Language Models},
  author={Hoffmann, Jordan and Borgeaud, Sebastian and Mensch, Arthur and Buchatskaya, Elena and Cai, Trevor and Rutherford, Eliza and Casas, Diego de Las and Hendricks, Lisa Anne and Welbl, Johannes and Clark, Aidan and others},
  journal={arXiv preprint arXiv:2203.15556},
  year={2022}
}

@article{radford2019language,
  title={Language Models are Unsupervised Multitask Learners},
  author={Radford, Alec and Wu, Jeffrey and Child, Rewon and Luan, David and Amodei, Dario and Sutskever, Ilya},
  journal={OpenAI technical report},
  year={2019}
}

@inproceedings{brown2020language,
  title={Language Models are Few-Shot Learners},
  author={Brown, Tom B. and Mann, Benjamin and Ryder, Nick and Subbiah, Melanie and Kaplan, Jared and Dhariwal, Prafulla and Neelakantan, Arvind and Shyam, Pranav and Sastry, Girish and Askell, Amanda and others},
  booktitle={Advances in Neural Information Processing Systems},
  volume={33},
  pages={1877--1901},
  year={2020}
}

@inproceedings{cook1971,
  author    = {Stephen A. Cook},
  title     = {The Complexity of Theorem-Proving Procedures},
  booktitle = {Proceedings of the 3rd Annual {ACM} Symposium on Theory of Computing ({STOC})},
  pages     = {151--158},
  year      = {1971},
}

@inproceedings{dao2019kernel,
  title = 	 {A Kernel Theory of Modern Data Augmentation},
  author =       {Dao, Tri and Gu, Albert and Ratner, Alexander and Smith, Virginia and De Sa, Chris and Re, Christopher},
  booktitle = 	 {Proceedings of the 36th International Conference on Machine Learning},
  pages = 	 {1528--1537},
  year = 	 {2019},
  editor = 	 {Chaudhuri, Kamalika and Salakhutdinov, Ruslan},
  volume = 	 {97},
  series = 	 {Proceedings of Machine Learning Research},
  month = 	 {09--15 Jun},
  publisher =    {PMLR},
}

@inproceedings{cubuk2020randaugment,
  title={Randaugment: Practical automated data augmentation with a reduced search space},
  author={Cubuk, Ekin D and Zoph, Barret and Shlens, Jonathon and Le, Quoc V},
  booktitle={Proceedings of the IEEE/CVF conference on computer vision and pattern recognition},
  pages={3008--3017},
  year={2020}
}

@article{ding2022data,
  title={Data augmentation for deep graph learning: A survey},
  author={Ding, Kaize and Xu, Zhe and Tong, Hanghang and Liu, Huan},
  journal={ACM SIGKDD Explorations Newsletter},
  volume={24},
  number={2},
  pages={61--77},
  year={2022},
  publisher={ACM New York, NY, USA}
}

@article{zhao2022graph,
  title={Graph data augmentation for graph machine learning: A survey},
  author={Zhao, Tong and Jin, Wei and Liu, Yozen and Wang, Yingheng and Liu, Gang and G{\"u}nnemann, Stephan and Shah, Neil and Jiang, Meng},
  journal={arXiv preprint arXiv:2202.08871},
  year={2022}
}

@inproceedings{rong2020dropedge,
  title={DropEdge: Towards Deep Graph Convolutional Networks on Node Classification},
  author={Rong, Yu and Huang, Wenbing and Xu, Tingyang and Huang, Junzhou},
  booktitle={International Conference on Learning Representations (ICLR)},
  year={2020}
}

@inproceedings{you2020graph,
  title={Graph contrastive learning with augmentations},
  author={You, Yuning and Chen, Tianlong and Sui, Yongduo and Chen, Ting and Wang, Zhangyang and Shen, Yang},
  booktitle={Advances in neural information processing systems},
  volume={33},
  pages={5812--5823},
  year={2020}
}

@article{shorten2019survey,
  title={A survey on image data augmentation for deep learning},
  author={Shorten, Connor and Khoshgoftaar, Taghi M},
  journal={Journal of Big Data},
  volume={6},
  number={1},
  pages={1--48},
  year={2019},
  publisher={Springer}
}

@inproceedings{selsam2019,
  author    = {Daniel Selsam and Matthew Lamm and Benedikt B{\"u}nz and Percy Liang and Leonardo de Moura and David L. Dill},
  title     = {Learning a {SAT} Solver from Single-Bit Supervision},
  booktitle = {International Conference on Learning Representations ({ICLR})},
  year      = {2019},
}

@article{li2024g4satbench,
  author  = {Zhaoyu Li and Jinpei Guo and Xujie Si},
  title   = {{G4SATBench}: Benchmarking and Advancing {SAT} Solving with Graph Neural Networks},
  journal = {Transactions on Machine Learning Research},
  year    = {2024},
}

@inproceedings{ozolins2022,
      title={Goal-aware neural SAT solver},
      author={Ozolins, Emils and Freivalds, Karlis and Draguns, Andis and Gaile, Eliza and Zakovskis, Ronalds and Kozlovics, Sergejs},
      booktitle={2022 International joint conference on neural networks (IJCNN)},
      pages={1--8},
      year={2022},
      organization={IEEE}

}

@article{cardillo2025,
  author  = {Franco Alberto Cardillo and Hamza Khyari and Umberto Straccia},
  title   = {{MILP-SAT-GNN}: Yet Another Neural {SAT} Solver},
  journal = {arXiv preprint arXiv:2507.01825},
  year    = {2025},
}

@article{eliasof2024,
      title={Quadratic Binary Optimization with Graph Neural Networks},
      author={Eliasof, Moshe and Haber, Eldad},
      journal={arXiv preprint arXiv:2404.04874},
      year={2024}
}

@inproceedings{gasse2019,
  author    = {Maxime Gasse and Didier Ch{\'e}telat and Nicola Ferroni and Laurent Charlin and Andrea Lodi},
  title     = {Exact Combinatorial Optimization with Graph Convolutional Neural Networks},
  booktitle = {Advances in Neural Information Processing Systems ({NeurIPS})},
  year      = {2019},
}

@inproceedings{ansotegui2012,
  author    = {Carlos Ans{\'o}tegui and Jes{\'u}s Gir{\'a}ldez-Cru and Jordi Levy},
  title     = {The Community Structure of {SAT} Formulas},
  booktitle = {Theory and Applications of Satisfiability Testing ({SAT})},
  pages     = {410--423},
  year      = {2012},
}

@inproceedings{cheng2024satgl,
  title={SATGL: An Open-Source Graph Learning Toolkit for Boolean Satisfiability},
  author={Cheng, HongTao and Liu, Jiawei and Zhai, Jianwang and Zhao, Mingyu and Yang, Cheng and Shi, Chuan},
  booktitle={2024 2nd International Symposium of Electronics Design Automation (ISEDA)},
  pages={746--751},
  year={2024},
  organization={IEEE}
}

@inproceedings{fu2025structuresat,
  title={Structure based SAT dataset for analysing GNN generalisation},
  author={Fu, Yi and Tompkins, Anthony and Song, Yang and Pagnucco, Maurice},
  journal={arXiv preprint arXiv:2502.11410},
  year={2025}
}

@inproceedings{peltonen2026,
  author    = {Saku Peltonen and Roger Wattenhofer},
  title     = {On the Expressive Power of {GNNs} for Boolean Satisfiability},
  booktitle = {International Conference on Learning Representations ({ICLR})},
  year      = {2026},
}

@inproceedings{khalil2022mipgnn,
  author    = {Elias B. Khalil and Christopher Morris and Andrea Lodi},
  title     = {{MIP-GNN}: A Data-Driven Framework for Guiding Combinatorial Solvers},
  booktitle = {Proceedings of the {AAAI} Conference on Artificial Intelligence},
  year      = {2022},
}

@inproceedings{neuroback2024,
  title={Neuroback: Improving CDCL SAT solving using graph neural networks},
  author={Wang, Wenxi and Hu, Yang and Tiwari, Mohit and Khurshid, Sarfraz and McMillan, Kenneth and Miikkulainen, Risto},
  journal={arXiv preprint arXiv:2110.14053},
  year={2021}
}

@inproceedings{yolcu2019,
  author    = {Emre Yolcu and Barnab{\'a}s P{\'o}czos},
  title     = {Learning Local Search Heuristics for Boolean Satisfiability},
  booktitle = {Advances in Neural Information Processing Systems ({NeurIPS})},
  year      = {2019},
}

@article{selfsatisfied2024,
  title={Self-Satisfied: An end-to-end framework for SAT generation and prediction},
  author={Serrano, Christopher R and Gallagher, Jonathan and Yamada, Kenji and Kopylov, Alexei and Warren, Michael A},
  journal={arXiv preprint arXiv:2410.14888},
  year={2024}
}

@article{achlioptas2005,
  author  = {Dimitris Achlioptas and Haixia Jia and Cristopher Moore},
  title   = {Hiding Satisfying Assignments: Two are Better than One},
  journal = {Journal of Artificial Intelligence Research},
  volume  = {24},
  pages   = {623--639},
  year    = {2005},
}

@misc{heule2016drattrim,
    author  = {Heule, Marijn J. H.},
      title   = {The DRAT format and DRAT-trim checker},
      journal = {arXiv},
      year    = {2016},
      doi     = {10.48550/arxiv.1610.06229}
}

\clearpage
\newpage

%%%%%%%%%%%%%%%%%%%%%%%%%%%%%%%%%%%%%%%%%%%%%%%%%%%%%%%%%%%%

\appendix

\crefalias{section}{appendix} % <--- Add this line!

\section{Additional Related Work}
\label{app:related_work}

% \paragraph{Graph representations for SAT.}
% A central modeling decision in SAT learning is how a CNF formula is represented as a graph. Bipartite constructions dominate the recent literature. NeuroSAT \citep{selsam2019} and G4SATBench represent formulas using literal--clause graphs, whereas MILP-SAT-GNN \citep{cardillo2025} builds a variable-constraint graph from a mixed-integer formulation. These encodings preserve clause structure and expose the incidence pattern between variables and constraints, but they also imply that variables co-occurring in a clause remain separated by at least two hops in the message-passing graph. Alternative non-bipartite constructions include the variable interaction graph (VIG) and related variants \citep{ansotegui2012}. Empirical comparisons, however, generally favor bipartite encodings for satisfiability prediction: SATGL \citep{cheng2024satgl} reports a clear advantage for literal--clause graphs, and \citet{fu2025structuresat} similarly find non-bipartite structures to be substantially weaker for this task.

\paragraph{Optimization-aware signals in graph learning.}
A related line of work enriches graph neural networks with optimization-derived information. BPGNN \citep{eliasof2024} incorporates a Quadratic Binary Optimization (QUBO) residual of the form $\mathbf{r}=\mathbf{h}\odot(A\mathbf{h}+\mathbf{b})$, thereby coupling message passing to an energy-based feasibility signal. MIP-GNN \citep{khalil2022mipgnn} propagates linear-constraint residual information to support branching decisions in mixed-integer optimization, and related learning-to-branch approaches use LP slack information as node features \citep{gasse2019}. Our setting differs from these works in both problem formulation and learning objective. We study SAT classification under a CNF-native ILP encoding and use the resulting residual quantities to enrich the graph representation itself, rather than to imitate branching or other solver decisions.

\paragraph{Training data generation and supervision.}
Most supervised SAT-GNN pipelines rely on solver-generated labels, which ties dataset construction directly to the runtime of a complete SAT solver. The standard pipeline generates a random $k$-CNF formula at the phase transition, calls a CDCL solver (CaDiCaL \citep{biere2020cadical} in G4SATBench, Glucose or MiniSAT in earlier work) to determine its label, and retains only formulas with the desired label, discarding the rest. This \emph{generate-and-label} loop scales poorly: CaDiCaL is a state-of-the-art conflict-driven clause-learning solver, yet its runtime still grows exponentially near the phase transition, making each labeled instance progressively more expensive as $n$ grows (see \Cref{tab:gen-efficiency}). DRAT-trim \citep{heule2016drattrim} provides additional UNSAT-core supervision on top of these labels. As noted by \citep{selfsatisfied2024}, once formulas reach moderate size, solver labeling can dominate the overall cost of the learning pipeline.
Several alternative generation strategies have been proposed. Graph generative models such as G2SAT \citep{you2019g2sat} learn to produce formulas with industrial-like structure by operating on literal-incidence graphs, but they require a pretrained generative model and still call a solver to verify labels. The SR generator used in NeuroSAT \citep{selsam2019} iteratively adds clauses and checks satisfiability after each addition, also solver-dependent. Planted-SAT constructions \citep{achlioptas2005} are solver-free but can produce unrealistic distributions. These observations point to a deeper challenge: the effectiveness of a generation strategy depends not only on correctness and speed but on the distributional alignment between generated and target instances. Our generation strategy is explicitly CNF-native and solver-free: the generated object is always a SAT or UNSAT formula with a correctness certificate by construction. Crucially, our target-aware variant matches the structural statistics of the downstream benchmark, which we show is the key factor determining augmentation quality.

% \textbf{Data augmentation in machine learning.} Beyond SAT-specific formula generation, our work connects to the broader paradigm of data augmentation, which is a cornerstone of robust representation learning \citep{shorten2019survey, cubuk2020randaugment}. While continuous domains such as computer vision readily admit semantics-preserving perturbations (e.g., cropping, rotation), augmenting discrete, graph-structured data is inherently more challenging \citep{ding2022data, zhao2022graph}. Graph data augmentation techniques often rely on random node dropping, feature masking, or edge perturbation \citep{rong2020dropedge, you2020graph}. However, applying these generic perturbations to combinatorial problems like SAT is highly destructive, as altering a single edge can inadvertently change the satisfiability status of the underlying formula. Consequently, augmentation for NP-hard problems requires structure-preserving techniques. Our framework bypasses destructive perturbations entirely by generating valid, benchmark-aligned formulas from the ground up, providing a domain-native augmentation strategy that scales combinatorial learning safely.

\section{Simple Algorithm Example}
\label{app:set_unsat_example}

Below we present a concrete example to demonstrate the target-aware formula generation procedures outlined in \Cref{alg:target-aware-sat,alg:ta-unsat}.

\subsection{SAT Data Generation}
We construct a satisfiable formula $\varphi_{\text{SAT}}$ of $m=2$ clauses over $n=3$ variables $x_1, x_2, x_3$. First, we sample a random reference assignment $x^* = \{x_1=1, x_2=0, x_3=1\}$. For the first clause $C_1$, suppose we sample width $k_1=3$ over the variables $\{x_1, x_2, x_3\}$ and a target slack $s_1^* = 1$. This target requires exactly $s_1^* + 1 = 2$ literals to evaluate to True under $x^*$, and the remaining $k_1 - s_1^* - 1 = 1$ literal to evaluate to False. To satisfy this, we select the positive literal $x_1$ and the negative literal $\neg x_2$ (both of which are True under $x^*$), alongside the negative literal $\neg x_3$ (which is False under $x^*$), yielding $C_1 = x_1 \vee \neg x_2 \vee \neg x_3$. 

For the second clause $C_2$, assume we sample width $k_2=2$ over $\{x_2, x_3\}$ with a target slack $s_2^* = 0$. This requires exactly one True literal and one False literal. Selecting the positive literal $x_3$ (True) and the positive literal $x_2$ (False) gives $C_2 = x_2 \vee x_3$. The resulting correctly labeled, slack-matched formula is therefore:
\begin{equation}
\varphi_{\text{SAT}} = (x_1 \vee \neg x_2 \vee \neg x_3) \wedge (x_2 \vee x_3).
\end{equation}

\subsection{UNSAT Data Generation}
To generate an unsatisfiable formula $\varphi_{\text{UNSAT}}$, the procedure embeds a resolution-tree core and pads it with slack-matched filler clauses. For the core construction, we select a dominant width $w=2$ and instantiate two variables, $x_1$ and $x_2$. We then generate all $2^w = 4$ possible polarity combinations to form a guaranteed contradiction:
\begin{equation}
C_{\text{core}} = (x_1 \vee x_2) \wedge (x_1 \vee \neg x_2) \wedge (\neg x_1 \vee x_2) \wedge (\neg x_1 \vee \neg x_2).
\end{equation}

Next, we generate additional filler clauses using the planted-slack method to align the formula with target benchmark statistics. We refer to the initial assignment for the remaining variables, for example  $x^* _3=1, x^*_4=0$. Suppose the algorithm samples a filler clause $C_5$ of width $k=2$ with target slack $s^* = 0$ over the variables $\{x_3, x_4\}$. This requires exactly one True literal and one False literal under $x^*$. Using the positive literal $x_3$ (which evaluates to True) and the positive literal $x_4$ (which evaluates to False) yields the filler clause $C_5 = x_3 \vee x_4$. 

Appending the filler to the core guarantees that the global formula remains strictly UNSAT while maintaining the desired structural properties:
\begin{equation}
\varphi_{\text{UNSAT}} = \underbrace{(x_1 \vee x_2) \wedge (x_1 \vee \neg x_2) \wedge (\neg x_1 \vee x_2) \wedge (\neg x_1 \vee \neg x_2)}_{\text{embedded core}} \wedge \underbrace{(x_3 \vee x_4)}_{\text{filler}}.
\end{equation}

\section{Generation Cost Scaling Visualization}
\label{app:generation_cost_plot}

\begin{figure}[h]
    \centering
    \includegraphics[width=0.8\textwidth]{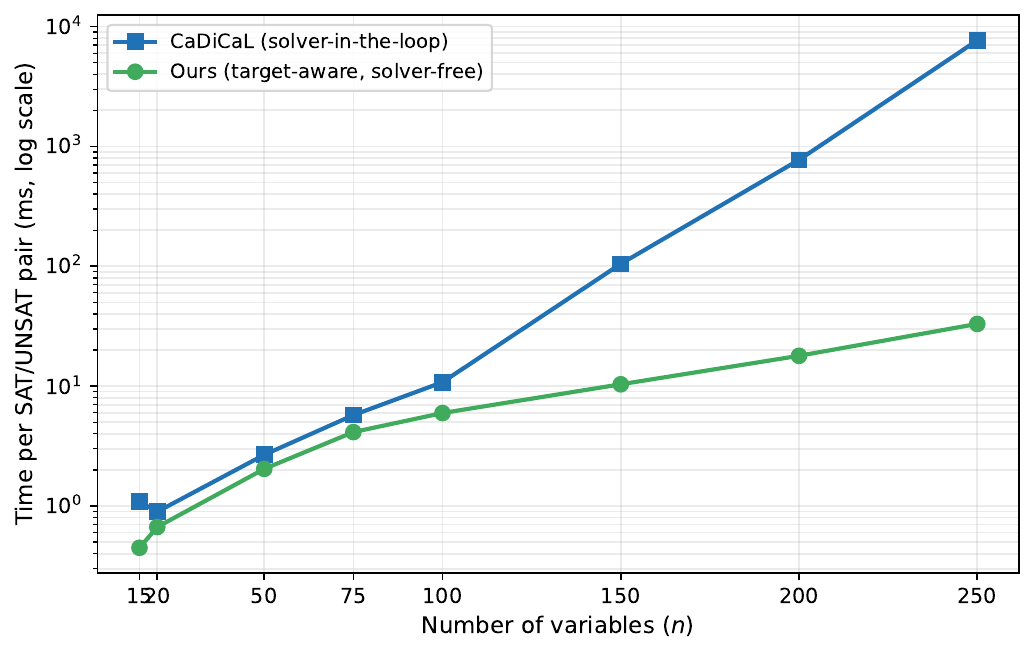}
    \caption{Generation cost vs. scale. The plot visualizes the data from Table~\ref{tab:gen-efficiency}, demonstrating the exponential scaling of the CaDiCaL-based baseline compared to the near-linear scaling of the proposed target-aware generator.}
    \label{fig:gen_cost_plot_app}
\end{figure}

Table~\ref{tab:gen-efficiency} provides the exact median wall-clock times required to generate correctly labeled SAT/UNSAT pairs, Figure~\ref{fig:gen_cost_plot_app} visualizes these scaling trajectories. The plot clearly illustrates the exponential divergence in computational cost between the standard solver-in-the-loop baseline (CaDiCaL) and our near-linear, solver-free generation approach as the problem size increases.

\section{Additional Results}
\label{app:additional_results}

\subsection{Scaling Behavior on Easy 3-SAT Instances}
\label{app:scaling_easy}

\begin{figure}[ht]
    \centering
    
    \begin{subfigure}[b]{0.48\textwidth}
        \centering
        \includegraphics[width=\textwidth]{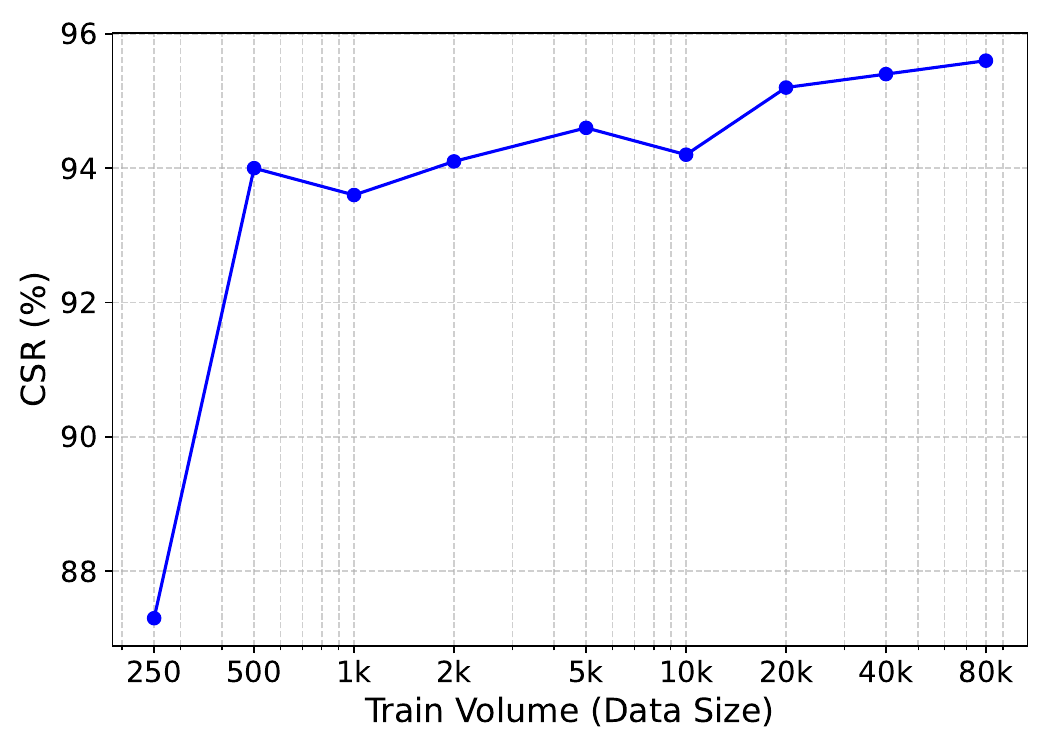}
        \caption{CSR.}
        \label{fig:csr_16l_easy}
    \end{subfigure}
    \hfill
    \begin{subfigure}[b]{0.48\textwidth}
        \centering
        \includegraphics[width=\textwidth]{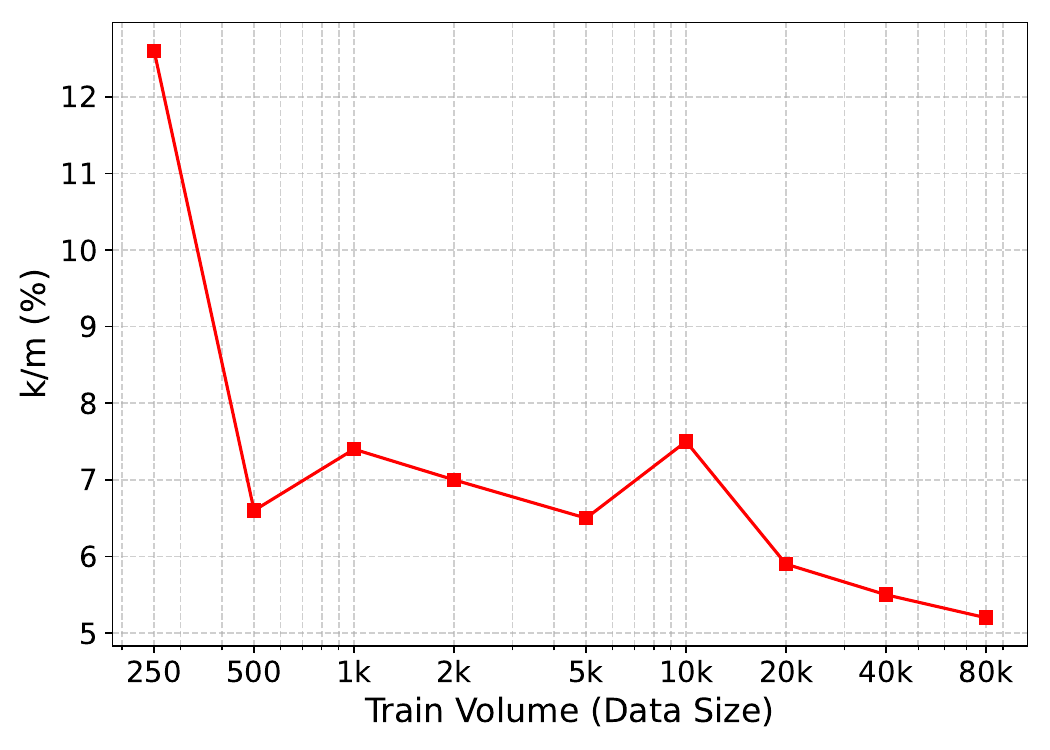}
        \caption{$k/m$ ratio.}
        \label{fig:km_16l_easy}
    \end{subfigure}
    
    \caption{Scaling behavior of the LPGNN trained on generated data as a function of training data volume. Evaluated on the G4SATBench easy 3-SAT test set. As the volume of training data increases, the Constraint Satisfaction Rate (CSR) consistently improves, indicating higher-quality assignment predictions for SAT instances. Concurrently, the $k/m$ ratio decreases, demonstrating that the predicted assignments for UNSAT instances violate progressively fewer clauses.}
    \label{fig:scaling_16l_easy_combined}
\end{figure}

In \Cref{sec:experiments}, we presented the scaling behavior of the LPGNN model evaluated on the Medium 3-SAT test set to demonstrate the impact of training data volume on assignment quality. We provide the corresponding scaling trajectories evaluated on the Easy 3-SAT test set in  \Cref{fig:scaling_16l_easy_combined}.

\subsection{Training Using the Original Benchmark}

To further evaluate the practical utility of our target-aware synthetic data as an augmentation strategy, we conducted experiments mixing generated instances with the existing G4SATBench training set. Notably, unlike our synthetic data, which features planted minimal-violation assignments by construction, the original benchmark does not provide corresponding assignment labels for UNSAT instances. Table \ref{tab:mixed-scaling-summary} summarizes the assignment quality of the model when trained exclusively on the benchmark data versus when augmented with an additional $40k$ synthetic instances. Evaluated on both the Easy and Medium 3-SAT test sets, the results demonstrate that injecting target-aware synthetic data effectively improves downstream assignment prediction.

\begin{table}[ht]
\centering
\caption{Scaling comparison of assignment quality for LPGNN trained on G4SATBench train set and generated synthetic target-aware data. Evaluated on the Easy and Medium 3-SAT test sets.}
\label{tab:mixed-scaling-summary}
\begin{tabular}{lcccc}
\toprule
 & \multicolumn{2}{c}{\textbf{Easy 3-SAT}} & \multicolumn{2}{c}{\textbf{Medium 3-SAT}} \\
\cmidrule(lr){2-3}\cmidrule(lr){4-5}
\textbf{Added Vol} & \textbf{CSR (\%)} $\uparrow$ & $\boldsymbol{k/m}$ \textbf{(\%)} $\downarrow$ & \textbf{CSR (\%)} $\uparrow$ & $\boldsymbol{k/m}$ \textbf{(\%)} $\downarrow$ \\
\midrule
G4SATBench   & $94.3$ & $6.6$ & $94.2$ & $6.3$ \\
G4SATBench + synthetic data & $95.1$ & $6.0$ & $95.1$ & $5.6$ \\
\bottomrule
\end{tabular}
\end{table}

% \begin{table}[ht]
% \centering
% \caption{Scaling comparison of assignment quality for LPGNN trained on G4SATBench train set plus generated data. Evaluated on the Easy and Medium 3-SAT test sets. Showing the benchmark-only baseline ($+0$) against $+40k$ generated instances. 3 seeds $\pm$ std.}
% \label{tab:mixed-scaling-summary}
% \begin{tabular}{lcccc}
% \toprule
%  & \multicolumn{2}{c}{\textbf{Easy 3-SAT}} & \multicolumn{2}{c}{\textbf{Medium 3-SAT}} \\
% \cmidrule(lr){2-3}\cmidrule(lr){4-5}
% \textbf{Added Vol} & \textbf{CSR (\%)} $\uparrow$ & $\boldsymbol{k/m}$ \textbf{(\%)} $\downarrow$ & \textbf{CSR (\%)} $\uparrow$ & $\boldsymbol{k/m}$ \textbf{(\%)} $\downarrow$ \\
% \midrule
% $+0$   & $94.3{\pm}0.1$ & $6.6{\pm}0.1$ & $94.2{\pm}0.4$ & $6.3{\pm}0.2$ \\
% $+40k$ & $95.1{\pm}0.3$ & $6.0{\pm}0.6$ & $95.1{\pm}0.2$ & $5.6{\pm}0.2$ \\
% \bottomrule
% \end{tabular}
% \end{table}

\subsection{Comparison With Generic Data}

We compare the model's scaling behavior under two data regimes. \emph{Generic data} refers to standard random $k$-CNF generation: SAT instances are produced by planting a random satisfying assignment and sampling clauses uniformly, while UNSAT instances are constructed via a backward resolution tree with random filler clauses. Neither generator is designed to match the LP residual distribution of the target benchmark.

\begin{figure}[ht]
\centering
\includegraphics[width=0.5\linewidth]{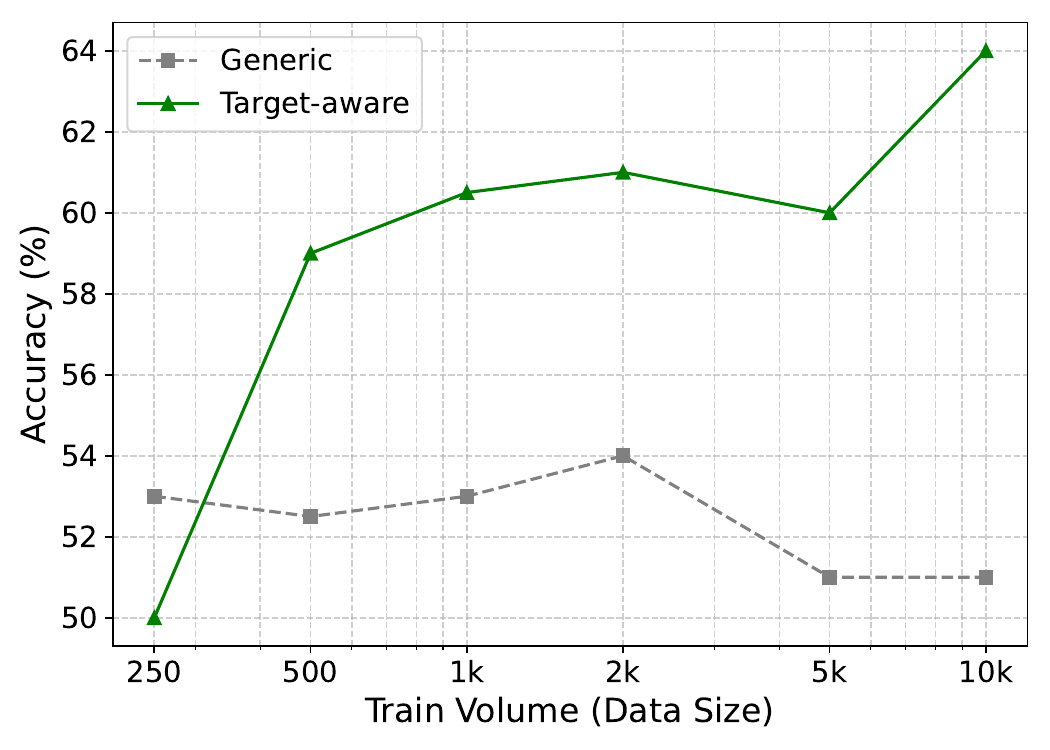}
\caption{Comparison of Accuracy (\%) scaling for LPGNN on the Easy 3-SAT test set. We show performance using a generic data baseline and the proposed target-aware synthetic data method. While using target-aware generated data shows increasing performance, using generic data remains near-random.}
\label{fig:ba_scaling_comparison}
\end{figure}

Training on generic data yields a flat performance profile — accuracy hovers just above random across all training volumes, indicating that the model fails to extract useful structure from misaligned data. Target-aware data, by contrast, drives a steady improvement from near-random at small volumes to a meaningful predictive signal at larger ones, as illustrated in~\Cref{fig:ba_scaling_comparison}. This gap highlights the importance of data alignment: the same architecture and loss function produce qualitatively different scaling behavior depending on the structural properties of the training data.

\subsection{LPGNN Generalization Tests}
\label{app:ood_lpgnn}

We provide results for generalization of the experiments discussed in Section~\ref{subsec:experiments_scaling_ood}. These experiments evaluate the robustness of the constraint satisfaction heuristics learned by the LPGNN model when tested on formula distributions that differ from the training set.

\begin{figure}[htbp]
    \centering
    % Row 1: CSR
    \begin{subfigure}[b]{0.48\textwidth}
        \centering
        \includegraphics[width=\textwidth]{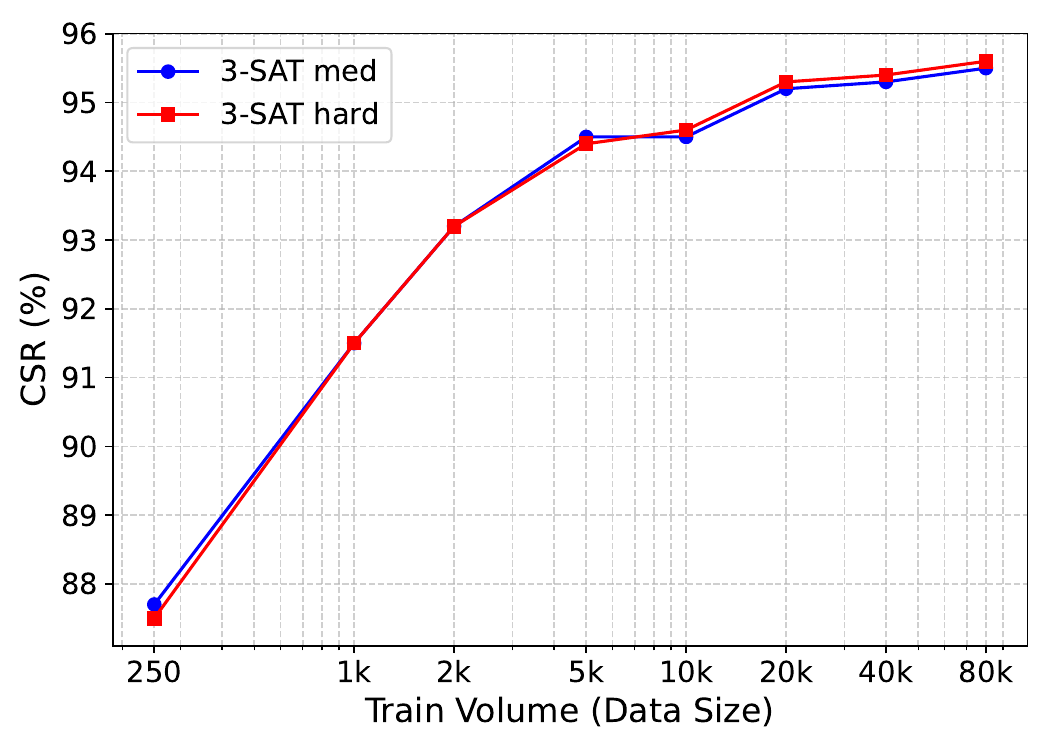}
        \caption{CSR: Evaluation on other 3-SAT families.}
        \label{fig:ood_easy_csr_3sat}
    \end{subfigure}
    \hfill
    \begin{subfigure}[b]{0.48\textwidth}
        \centering
        \includegraphics[width=\textwidth]{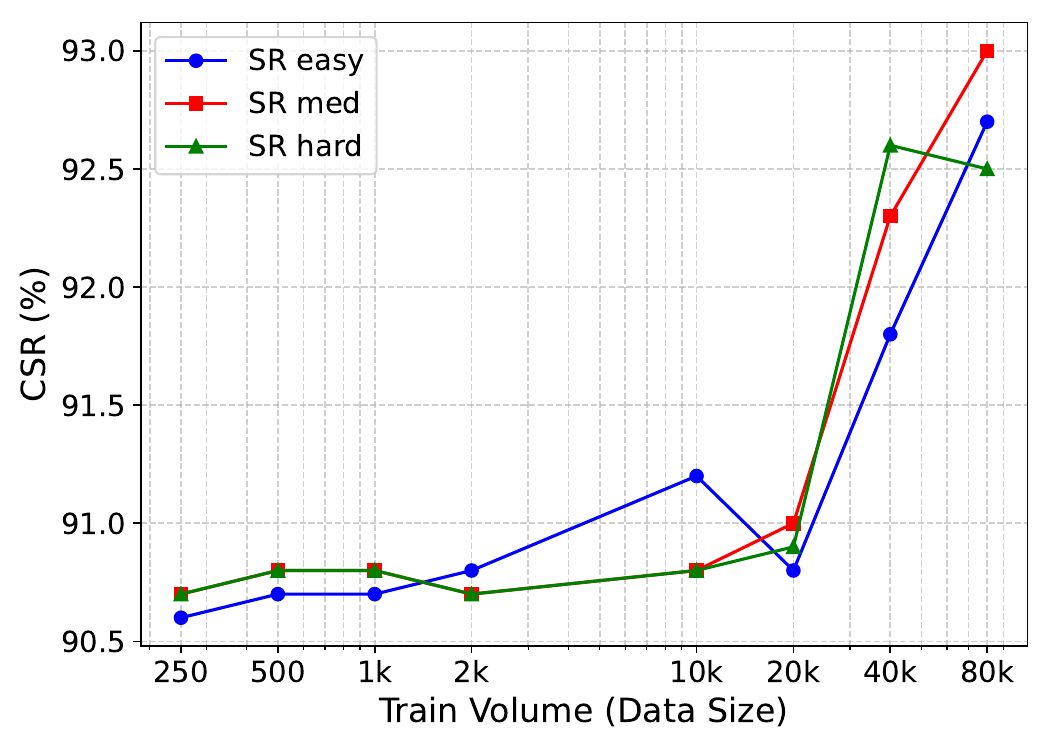}
        \caption{CSR: Evaluation on SR families.}
        \label{fig:ood_easy_csr_sr}
    \end{subfigure}
    
    \vspace{0.5cm} % Vertical space between rows
    
    % Row 2: k/m
    \begin{subfigure}[b]{0.48\textwidth}
        \centering
        \includegraphics[width=\textwidth]{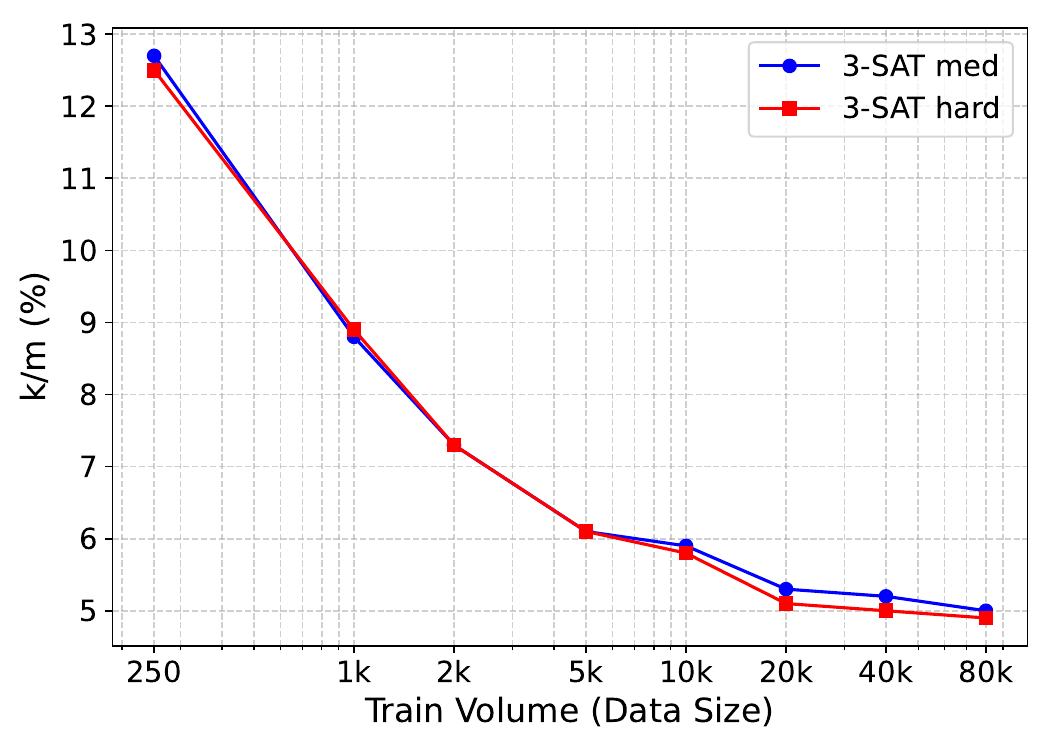}
        \caption{$k/m$: Evaluation on other 3-SAT families.}
        \label{fig:ood_easy_km_3sat}
    \end{subfigure}
    \hfill
    \begin{subfigure}[b]{0.48\textwidth}
        \centering
        \includegraphics[width=\textwidth]{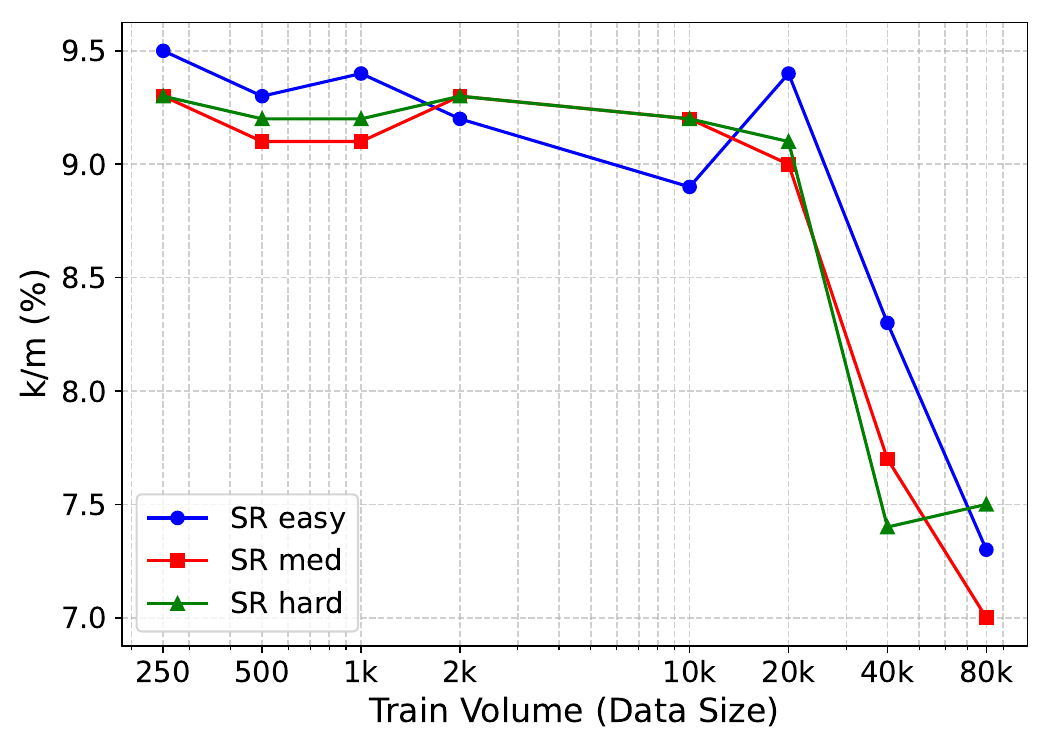}
        \caption{$k/m$: Evaluation on SR families.}
        \label{fig:ood_easy_km_sr}
    \end{subfigure}
    
    \caption{Additional tests results: Constraint Satisfaction Rate (CSR, \%) and violated-clause fraction ($k/m$, \%) for LPGNN trained on generated \textbf{easy 3-SAT}.}
    \label{fig:ood_easy_combined}
\end{figure}

Figure~\ref{fig:ood_easy_combined} illustrates the generalization trajectories for the model trained on the \emph{easy 3-SAT} generated instances. The top row displays the Constraint Satisfaction Rate (CSR), while the bottom row shows the violated-clause fraction ($k/m$). The left column evaluates zero-shot transfer to harder 3-SAT distributions (medium and hard), and the right column evaluates transfer to random structural (SR) formulas. We observe that performance on these structurally proximate families scales favorably with the training data volume, staying well above random-assignment baselines.

\begin{figure}[htbp]
    \centering
    % Row 1: CSR
    \begin{subfigure}[b]{0.48\textwidth}
        \centering
        \includegraphics[width=\textwidth]{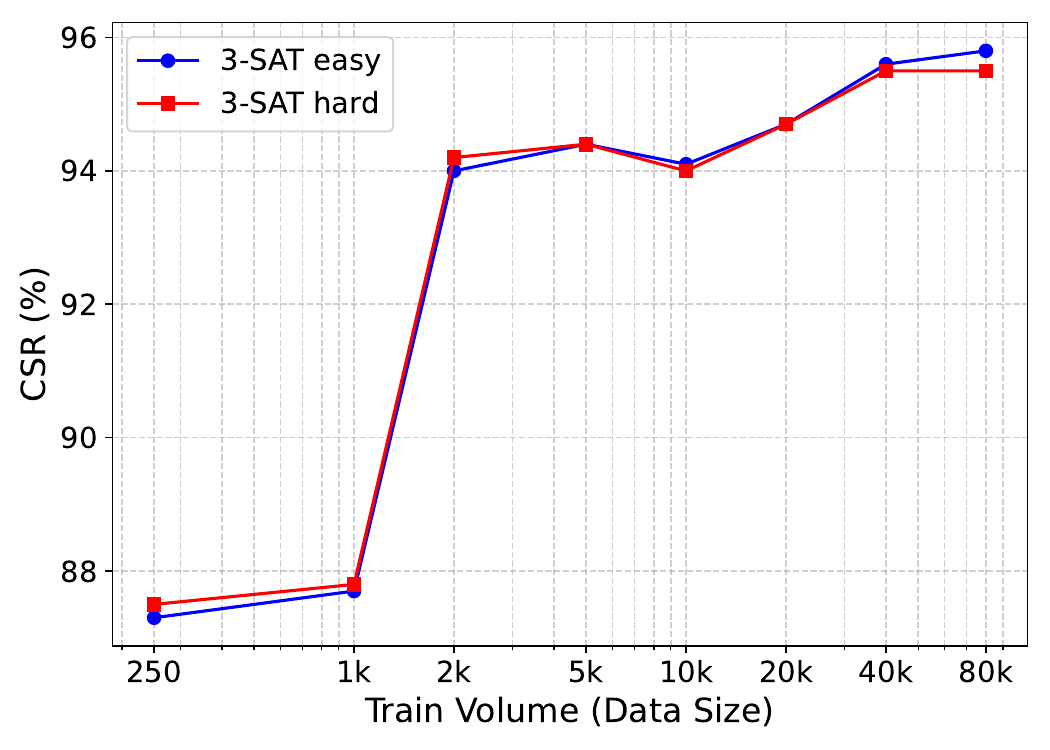}
        \caption{CSR: Evaluation on other 3-SAT families.}
        \label{fig:ood_med_csr_3sat}
    \end{subfigure}
    \hfill
    \begin{subfigure}[b]{0.48\textwidth}
        \centering
        \includegraphics[width=\textwidth]{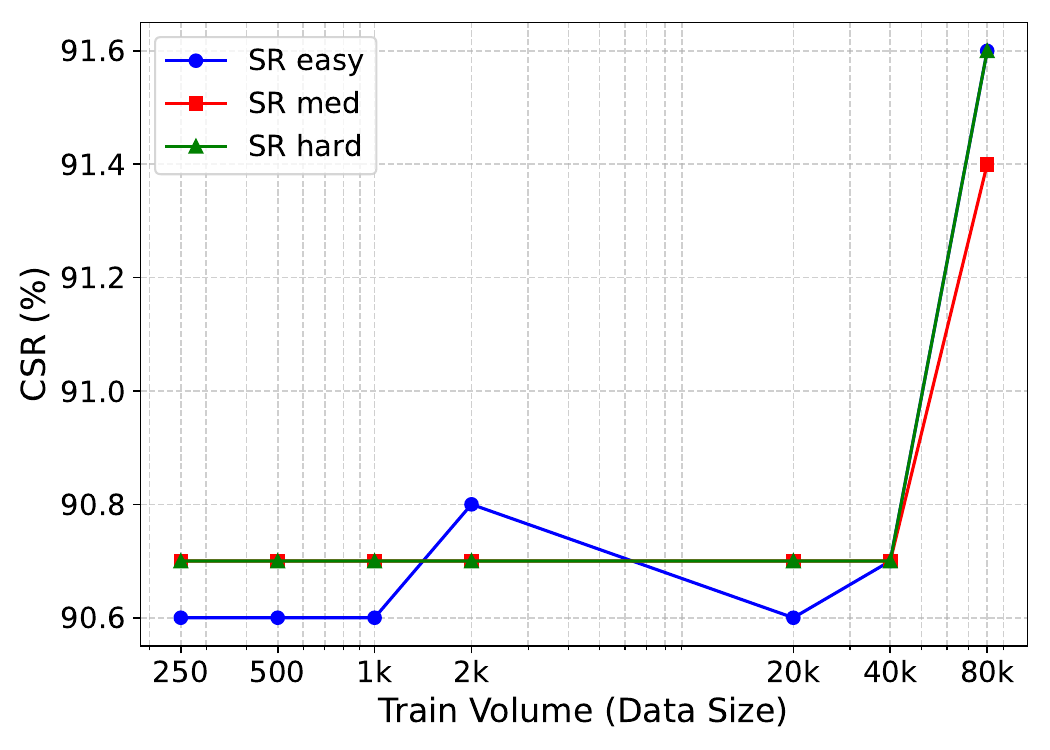}
        \caption{CSR: Evaluation on SR families.}
        \label{fig:ood_med_csr_sr}
    \end{subfigure}
    
    \vspace{0.5cm} % Vertical space between rows
    
    % Row 2: k/m
    \begin{subfigure}[b]{0.48\textwidth}
        \centering
        \includegraphics[width=\textwidth]{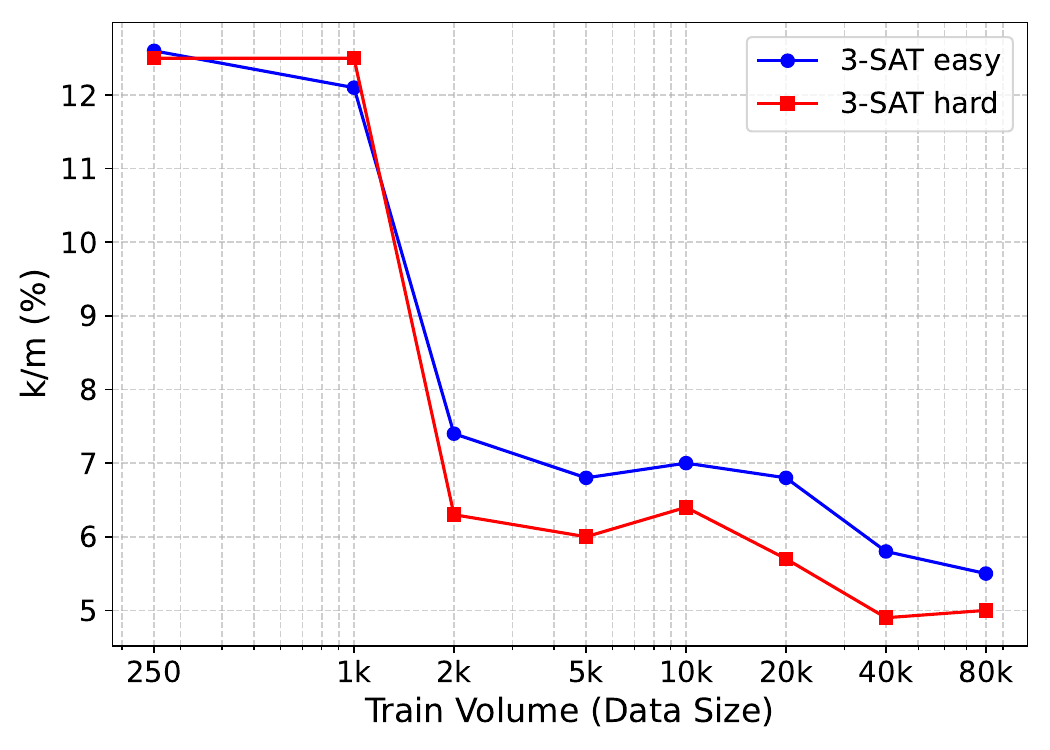}
        \caption{$k/m$: Evaluation on other 3-SAT families.}
        \label{fig:ood_med_km_3sat}
    \end{subfigure}
    \hfill
    \begin{subfigure}[b]{0.48\textwidth}
        \centering
        \includegraphics[width=\textwidth]{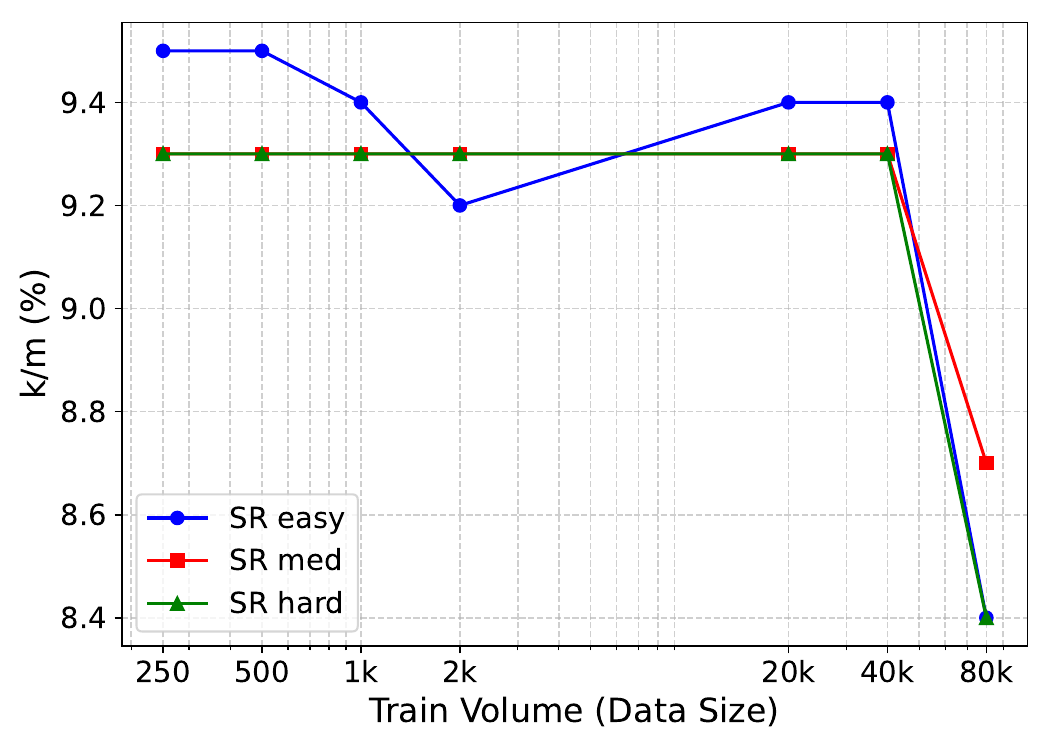}
        \caption{$k/m$: Evaluation on SR families.}
        \label{fig:ood_med_km_sr}
    \end{subfigure}
    
    \caption{Additional tests results: Constraint Satisfaction Rate (CSR, \%) and violated-clause fraction ($k/m$, \%) for LPGNN trained on generated \textbf{medium 3-SAT}.}
    \label{fig:ood_med_combined}
\end{figure}

Similarly, Figure~\ref{fig:ood_med_combined} presents the corresponding OOD results for the model trained on the \emph{medium 3-SAT} generated instances. The model is evaluated on the easy and hard 3-SAT families (left column) as well as the SR families (right column). These trajectories confirm that the combined training objective enables robust transfer within related combinatorial domains. The assignment quality consistently improves as the base model is exposed to a larger volume of target-aligned synthetic data.

% ----------------------------------------------------
% FIGURE 1: Easy 3-SAT (CSR and k/m)
% ----------------------------------------------------

% ----------------------------------------------------
% FIGURE 2: Medium 3-SAT (CSR and k/m)
% ----------------------------------------------------

\section{Ablation study}
\label{app:ablation_study}

To isolate the contribution of the linear programming (LP) residual mechanism, we conducted an ablation study comparing the full LPGNN architecture against a baseline model utilizing only the underlying GIN backbone. Table \ref{tab:ilp-ablation-10k} presents the performance of both models evaluated on the G4SATBench Easy 3-SAT test set, following training on 10k generated instances.
The results demonstrate that incorporating ILP residuals yields consistent improvements across all measured metrics. These findings suggest that injecting algebraic feasibility signals into the message-passing phase enhances the network's capacity for both satisfiability classification and assignment prediction.

\begin{table}[ht]
\centering
\caption{Comparison of the Full LPGNN architecture (with residuals) against the GIN backbone baseline on the Easy 3-SAT test set. The Full LPGNN demonstrates superior performance across all three metrics.}
\label{tab:ilp-ablation-10k}
\begin{tabular}{lcc}
\toprule
\textbf{Metric} & \textbf{Full LPGNN} & \textbf{Backbone} \\
\midrule
Accuracy (\%) $\uparrow$ & $64.0$ & $60.5$ \\
CSR (\%) $\uparrow$         & $94.2$ & $93.1$ \\
$k/m$ (\%) $\downarrow$     & $7.5$  & $8.2$ \\
\bottomrule
\end{tabular}
\end{table}

% =================================================================
\section{Experimental Setup}
\label{app:dataset_statistics}
% =================================================================

\subsection{Dataset Details}

We utilize the G4SATBench benchmark \citep{li2024g4satbench}, which provides random 3-SAT instances categorized into three difficulty levels, alongside additional formula families designed for out-of-distribution (OOD) evaluation. We note that the slack distribution of SAT and UNSAT instances described in \Cref{alg:target-aware-sat,alg:ta-unsat} refers to the training sets of the benchmark.

\paragraph{3-SAT Instances.} 
The core 3-SAT dataset is constructed near the theoretical satisfiability phase transition, exhibiting a clause-to-variable ratio of $\alpha = m/n \approx 4.27$. These instances are partitioned by difficulty into Easy, Medium, and Hard tiers, with each tier maintaining an equal balance of satisfiable and unsatisfiable formulas. Both the Easy and Medium tiers contain 1,600 training instances, 200 validation instances, and 200 test instances. The Easy instances range from $n=10$ to $40$ variables ($m \in [55, 175]$) with an average $\alpha$ of $4.59$. The Medium instances range from $n=40$ to $200$ variables ($m \in [175, 850]$) with an average $\alpha$ of $4.28$. The Hard tier is reserved exclusively for testing, comprising 200 instances with $n \in [200, 300]$ variables and $m \in [853, 1278]$ clauses, yielding an average $\alpha$ of $4.26$.

\paragraph{Additional Families.} 
To assess out-of-distribution generalization, we utilize the SR (random, 3-SAT-like) family provided by the benchmark. Consistent with the 3-SAT testing protocol, the SR evaluation sets contain 200 test instances (100 SAT and 100 UNSAT) for each of the three difficulty levels. For the satisfiable subsets, the SR instances exhibit a mean variable and clause count of $n=27$, $m=151$ for the Easy tier; $n=123$, $m=680$ for the Medium tier; and $n=302$, $m=1640$ for the Hard tier.

% =================================================================
\subsection{Implementation Details}
\label{app:experiments_detail}
% =================================================================
All of our runs are single-GPU on NVIDIA RTX~4090 (24\,GB). In the following, we elaborate the experimental details.

\subsubsection{LPGNN}
Our experiments use the LPGNN model with a GIN \citep{xu2019how} backbone, with per-layer
residual connections.  
Table~\ref{tab:lpgnn-config} lists the full configuration.

\begin{table}[ht]
\centering
\caption{LPGNN training configuration.}
\label{tab:lpgnn-config}
\begin{tabular}{l c}
\toprule
Hyperparameter & Value \\
\midrule
Hidden dimension $d$    & 128 \\
Layers         & 16 \\
% Parameters              & 856K \\
% Graph type              & LPGNN-lite ($d_\text{in}=7$) \\
Optimiser               & Adam \citep{kingma2015adam} \\
Learning rate           & $10^{-3}$ \\
Weight decay            & $10^{-6}$ \\
% Gradient clip (max norm)& 1.0 \\
Batch size (3-SAT easy / medium) & 256 / 64 \\
Max epochs              & 100 \\
Early-stopping patience & 15 \\
Loss & Classification Loss $+$ Binary Cross Entropy \\
Scheduler               & ReduceLROnPlateau (factor$=$0.5, patience$=$5) \\
% Loss weights $(\lambda_\text{cls},\,\lambda_\text{assign})$ & $(1.0,\;1.0)$ \\
% Hardware                & NVIDIA RTX 4090 (24\,GB) $\times$ 2 \\
\bottomrule
\end{tabular}
\end{table}

% =================================================================
\subsubsection{Additional Architecture}
\label{app:baseline-config}
% =================================================================

\paragraph{NLocalSAT~\citep{zhang2020nlocalsat}.}
NLocalSAT extends NeuroSAT~\citep{selsam2019} with a per-variable assignment
head that predicts a soft variable assignment at each inference step.  The
model operates on the Literal--Clause Graph (LCG): $2n$ literal nodes and $m$
clause nodes exchange messages via an LSTM recurrent cell for $T$ steps,
sharing complement information between positive and negative literals after
each iteration.  The per-variable assignment logits are derived by averaging
the positive and negative literal embeddings for each variable.

\paragraph{QuerySAT~\citep{ozolins2022}.}
QuerySAT likewise operates on the LCG with LSTM-based message passing, but introduces a \emph{query gradient mechanism}: at
each step the network produces a trial variable assignment, evaluates it
against a differentiable clause-loss (fraction of unsatisfied clauses), and
computes the gradient of that loss with respect to the trial assignment. This gradient vector is concatenated back into
the literal embeddings as additional input to the next LSTM step, giving the
network explicit first-order feedback about which variables are most
responsible for unsatisfied clauses. Full
configuration details are in Table~\ref{tab:baseline-config}.

Table~\ref{tab:baseline-config} summarises the configurations used for the
NLocalSAT and QuerySAT.  Both models operate on the
Literal--Clause Graph (LCG) and are trained with the same continuous-volume
pipeline used for LPGNN.

\begin{table}[ht]
\centering
\caption{Training configuration for NLocalSAT and QuerySAT.  All
hyperparameters not listed here match the LPGNN configuration in
Table~\ref{tab:lpgnn-config}.}
\label{tab:baseline-config}
\begin{tabular}{l c c}
\toprule
Hyperparameter & NLocalSAT & QuerySAT \\
\midrule
% Reference            & \citet{zhang2020nlocalsat} & \citet{ozolins2022} \\
% Graph representation & LCG & LCG \\
% Recurrent cell       & LSTM & LSTM \\
Message-passing steps & 16 & 32 \\
Hidden dimension & 128 & 128 \\
Query dimension      & ---  & 128 \\
% Parameters           & 462K & 363K \\
% Input features $d_\text{in}$ & 3 & 3 \\
Optimiser            & Adam \citep{kingma2015adam} & Adam \citep{kingma2015adam} \\
Learning rate        & $10^{-3}$ & $10^{-3}$ \\
Weight decay         & $10^{-6}$ & $10^{-6}$ \\
Batch size (easy / medium) & 256 / 64 & 256 / 64 \\
Max epochs           & 100 & 100 \\
Early-stopping patience & 15 & 15 \\
% Loss weights $(\lambda_\text{cls},\,\lambda_\text{assign})$ & $(1.0,\;1.0)$ & $(1.0,\;1.0)$ \\
\bottomrule
\end{tabular}
\end{table}

\section{Limitations}
\label{app:limitations}
While our target-aware framework provides a scalable source of labeled data, synthetic data augmentation inherently risks introducing distributional bias. Although our generator effectively matches the structural statistics, such as clause width and slack distributions, of a target benchmark, manufactured instances cannot perfectly replicate the complex exact logical dependencies and hidden community structures inherent in real-world or specialized industrial SAT instances. Consequently, there is a limitation common to synthetic data generation: models may inadvertently overfit to the artifacts or regularities of the generation algorithm rather than learning the true underlying combinatorial hardness. This synthetic-to-real gap implies that while augmented data serves as a powerful regularizer for in-distribution or structurally proximate instances, it does not guarantee robust generalization to entirely out-of-distribution families or highly structured industrial problems whose latent features are not captured by the target statistics. 
% We additionally discuss  broader societal impacts in \Cref{app:broader_impacts}.

\section{Broader Impacts}
\label{app:broader_impacts}
The ability to efficiently train scalable, learning-based heuristics for NP-hard problems carries dual-use implications. On the positive side, accelerating Boolean satisfiability directly benefits formal verification, software reliability, and hardware design, contributing to the development of safer and more robust computing systems. Furthermore, our solver-free data generation pipeline significantly reduces the computational bottleneck and energy consumption typically associated with producing large-scale training datasets for combinatorial problems. However, since SAT solvers are highly generalizable, advancements in this domain could theoretically be leveraged for adversarial purposes, such as cryptanalysis and breaking cryptographic protocols. We mitigate these concerns by focusing our evaluation on standard, open-source benchmarks and clearly present our data generation framework to promote transparent, defensive research.

%%%%%%%%%%%%%%%%%%%%%%%%%%%%%%%%%%%%%%%%%%%%%%%%%%%%%%%%%%%%

% \clearpage
% \newpage
% \input{checklist.tex}

\end{document}